\documentclass[10pt,twocolumn,letterpaper]{article}

\usepackage{iccv}
\usepackage{times}
\usepackage{epsfig}
\usepackage{graphicx}
\usepackage{amsmath}
\usepackage{amssymb}

\usepackage[letterspace=-120]{microtype}
\usepackage{algorithm}
\usepackage{algorithmic}
\usepackage{subcaption}
\usepackage[font=small]{caption}
\usepackage{dsfont}
\usepackage{xcolor,colortbl}
\usepackage{cancel}
\usepackage{nicefrac}
\usepackage{soul}

\definecolor{Gray}{gray}{0.85}
\definecolor{Gray2}{gray}{0.93}
\definecolor{yellowGreen}{rgb}{0.5,0.8,0.4}

\usepackage[pagebackref=true,breaklinks=true,letterpaper=true,colorlinks,bookmarks=false]{hyperref}

\iccvfinalcopy 

\def\iccvPaperID{1549} 

\ificcvfinal\fi
\begin{document}

\title{Learning to Divide and Conquer for Online Multi-Target Tracking}

\author{Francesco Solera\quad Simone Calderara\quad Rita Cucchiara\\\\
Department of Engineering\\
University of Modena and Reggio Emilia\\
{\tt\small name.surname@unimore.it}
}

\maketitle

\begin{abstract}
Online Multiple Target Tracking (MTT) is often addressed within the \emph{tracking-by-detection} paradigm. Detections are previously extracted independently in each frame and then objects trajectories are built by maximizing specifically designed coherence functions. Nevertheless, ambiguities arise in presence of occlusions or detection errors. 
In this paper we claim that the ambiguities in tracking could be solved by a selective use of the features, by working with more reliable features if possible and exploiting a deeper representation of the target only if necessary. To this end, we propose an online divide and conquer tracker for static camera scenes, which partitions the assignment problem in local subproblems and solves them by selectively choosing and combining the best features. The complete framework is cast as a structural learning task that unifies these phases and learns tracker parameters from examples.
Experiments on two different datasets highlights a significant improvement of tracking performances (MOTA +10\%) over the state of the art.  
%
\end{abstract}


\section{Introduction}
Multiple Target Tracking (MTT) is the task of extracting the continuous path of relevant objects across a set of subsequent frames. Due to the recent advances in object detection~\cite{dollar_fast_2014,Benenson2014Eccvw}, the problem of MTT is often addressed within the \emph{tracking-by-detection} paradigm. Detections are previously extracted independently in each frame and then objects trajectories are built by maximizing specifically designed coherence functions~\cite{milan_continuous_2014,berclaz_multiple_2011,possegger_occlusion_2014,bae_robust_2014,dicle_way_2013,wu_online_2013}.
Tracking objects through detections can mitigate drifting behaviors introduced by prediction steps but, on the other hand, it forces the tracker
to work in adverse conditions, due to the frequent occurrence of false and miss detections. 

The majority of approaches address MTT offline, \emph{i.e.}~by exploiting detections from a set of frames~\cite{milan_continuous_2014,berclaz_multiple_2011,dicle_way_2013} through global optimization. Offline methods benefit from the bigger portion of video sequence they dispose of to establish spatio-temporal coherence, but can not be used in real-time applications.
Conversely, online methods track the targets frame-by-frame; they have a larger spectra of application but must be both accurate and fast despite working with less data. In this context, the robustness of the features play a major role in the online MTT task.
Some approaches claim the adoption of complex targets models~\cite{bae_robust_2014,wu_online_2013} to be the solution, while others argue that this complexity may affect the long-term robustness~\cite{smeulders_visual_2014}. For instance, in large crowds people appearance is rarely informative.
As a consequence, tracking robustness is often achieved by focusing on spatial features~\cite{possegger_occlusion_2014}, finding them more reliable than visual ones.
\begin{figure}[t]
        \centering
        \includegraphics[width=1\columnwidth]{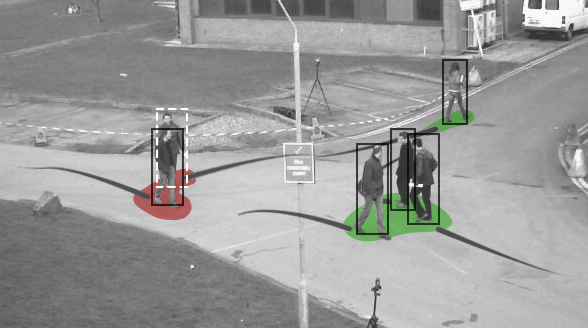}
        \caption{The scene is partitioned in local zones. Green zones is where the same number of tracks and detections are present. Red zones, where miss and false detections (white dashed contours) are discovered and solving the associations may call for complex appearance or motion features.}
        \label{fig:cover}
\end{figure}

We do believe that many of the ambiguities in tracking could be solved by a selective use of the features, by working with more reliable features if possible and exploiting a deeper representation of the target only if necessary. In fact, a simple spatial association is often sufficient while, as clutter or confusion arise, an improved association scheme on more complex features is needed (Fig.~\ref{fig:cover}).\\
\noindent In this paper a novel approach for online MTT in static camera scenes is proposed. The method selects the most suitable features to solve the frame-by-frame associations depending on the surrounding scene complexity.\\Specifically, our contributions are:
\begin{itemize}
	\item an online method based on Correlation Clustering that learns to \emph{divide} the global association task in smaller and localized association subproblems (Sec.~\ref{sec:divide}),
	\item a novel extension to the Hungarian association scheme, flexible enough to be applied to any set of preferred features and able to \emph{conquer} trivial and complex subproblems by selectively combining the features (Sec.~\ref{sec:conquer}),
\item an online Latent Structural SVM (LSSVM) framework to combine the \emph{divide} and \emph{conquer} steps and to learn from examples all the tracker parameters (Sec.~\ref{sec:learning}).

\end{itemize}
The algorithm works by alternating between (a) learning the affinity measure of the Correlation Clustering as a latent variable and (b) learning the optimal combinations for both simple and complex features to be used as cost functions by the Hungarian.
Results on public benchmarks underline a major improvement in tracking accuracy over current state of the art online trackers (+10\% MOTA).\\

\noindent The work takes inspiration from the human perceptive behavior, further introduced in Sec.~\ref{sec:human}. According to the widely accepted two-streams hypothesis by Goodale and Milner~\cite{Goodale199220}, the use of motion and appearance information is localized in the temporal lobe (\emph{what} pathway), while basic spatial cues are processed in the parietal lobe (\emph{where} pathway). This suggests our brain processes and exploits information in different and specific ways as well.

\section{Related works}
Tab. \ref{tab:related} reports an overview of recent tracking-by-detection literature approaches separating online and offline methods and indicates the adoption of tracklets (T), appearance models (A) and complex learning schemes (L).
Offline methods~\cite{berclaz_multiple_2011,milan_continuous_2014,hofmann_unified_2013,5206735} are out of the scope of the paper and are reported for the sake of completeness.\\
Tracklets are the results of an intermediate hierarchical association of the detections and are commonly used by both offline and online solutions ~\cite{5206735,hofmann_unified_2013,Yang_2014}. In these approaches, high confidence associations link detections in a pre-processing step and then optimization techniques are employed to link tracklets into trajectories. Nevertheless tracklets creation involves solving a frame by frame assignment problem by thresholding the final association cost and errors in tracklets affect the tracking results as well.    
\\
In addition, online methods often try to compensate the lack of spatiotemporal information through the use of appearance or other complex features model. Appearance model is typically handled by the adoption of a classifier for each tracked target~\cite{wu_online_2013} and data associations is often finalized through an averaged sum of classifiers scores, ~\cite{5459278,bae_robust_2014}. As a consequence, learning is on targets model, not on associations. 
Moreover, online methods also need to cope with drifting when updating their targets model. One possible solution is to avoid model updating when uncertainties are detected in the data, \emph{i.e.}~ a detection cannot be paired to a sufficiently close previous trajectory~\cite{bae_robust_2014}. Nevertheless, any error introduced into the model can rapidly lead to tracking drift and wrong appearance learning. Building on these considerations, Possegger~\emph{et~al.}~\cite{possegger_occlusion_2014} does not consider appearance at all and only work with distance and motion data.
\begin{table}[t]
\begin{tabular}{|l|c|c|c|c|c|c|}
\hline 
\multicolumn{2}{ |l| }{}   & C & A & T & L & M \\
\hline
\hline 
\multicolumn{7}{ |l| }{{\bf Offline methods}}\\
\hline 
Berclaz~\emph{et al.}~\cite{berclaz_multiple_2011} & 2011 & \checkmark & • & • & • & • \\ 
\hline 
Milan~\emph{et al.}~\cite{milan_continuous_2014} & 2014 & \checkmark & \checkmark & \checkmark & • & \checkmark \\ 
\hline 
Hoffman~\emph{et~al.}~\cite{hofmann_unified_2013} & 2013 & • & \checkmark & \checkmark & \checkmark & • \\ 
\hline 
Li~\emph{et~al.}~\cite{5206735} & 2009 & • & • & \checkmark & \checkmark & • \\
\hline 
\hline 
\multicolumn{7}{ |l| }{{\bf Online methods}}\\
\hline 
Yang and Nevatia~\cite{Yang_2014} & 2014 & • & • & \checkmark & \checkmark & • \\ 
\hline 
Breitenstein~\emph{et~al.}~\cite{5459278} & 2009 &  & \checkmark & • & • & • \\ 
\hline 
Bae and Yoon~\cite{bae_robust_2014} & 2014 & \checkmark & \checkmark & • & • & \checkmark \\ 
\hline 
Possegger~\emph{et~al.}~\cite{possegger_occlusion_2014} & 2014 & \checkmark & • & • & • & • \\ 
\hline
Wu~\emph{et~al.}~\cite{wu_online_2013} & 2013 & • & \checkmark & • & • & • \\ 
\hline
\hline 
{\bf Our proposal} & 2015 & \checkmark & \nicefrac{1}{2} & • & \checkmark & \checkmark \\ 
\hline 
\end{tabular}
\caption{Overview of offline and online related works in terms of code availability (C), appearance models (A), tracklets computation (T), associations learning (L) and presence in the MOT Challenge competition (M). In our method, use of appearance set to \nicefrac{1}{2} means only when needed.}
\label{tab:related}
\end{table}

\noindent Differently from the aforementioned online learning methods, our approach is not hierarchical and we do not compute intermediate tracklets because errors in the tracklets corrupt the learning data. Similarly to \cite{bae_robust_2014}, we model a score of uncertainty but based on distance information only and not on the target model, since distance can not drift over time. This enables us to invoke appearance and other less stable features only when truly needed as in the case of missing detections, occluded objects or new tracks.

\section{Related perception studies}
\label{sec:human}
\begin{figure}[t]
        \centering
        \includegraphics[width=\columnwidth]{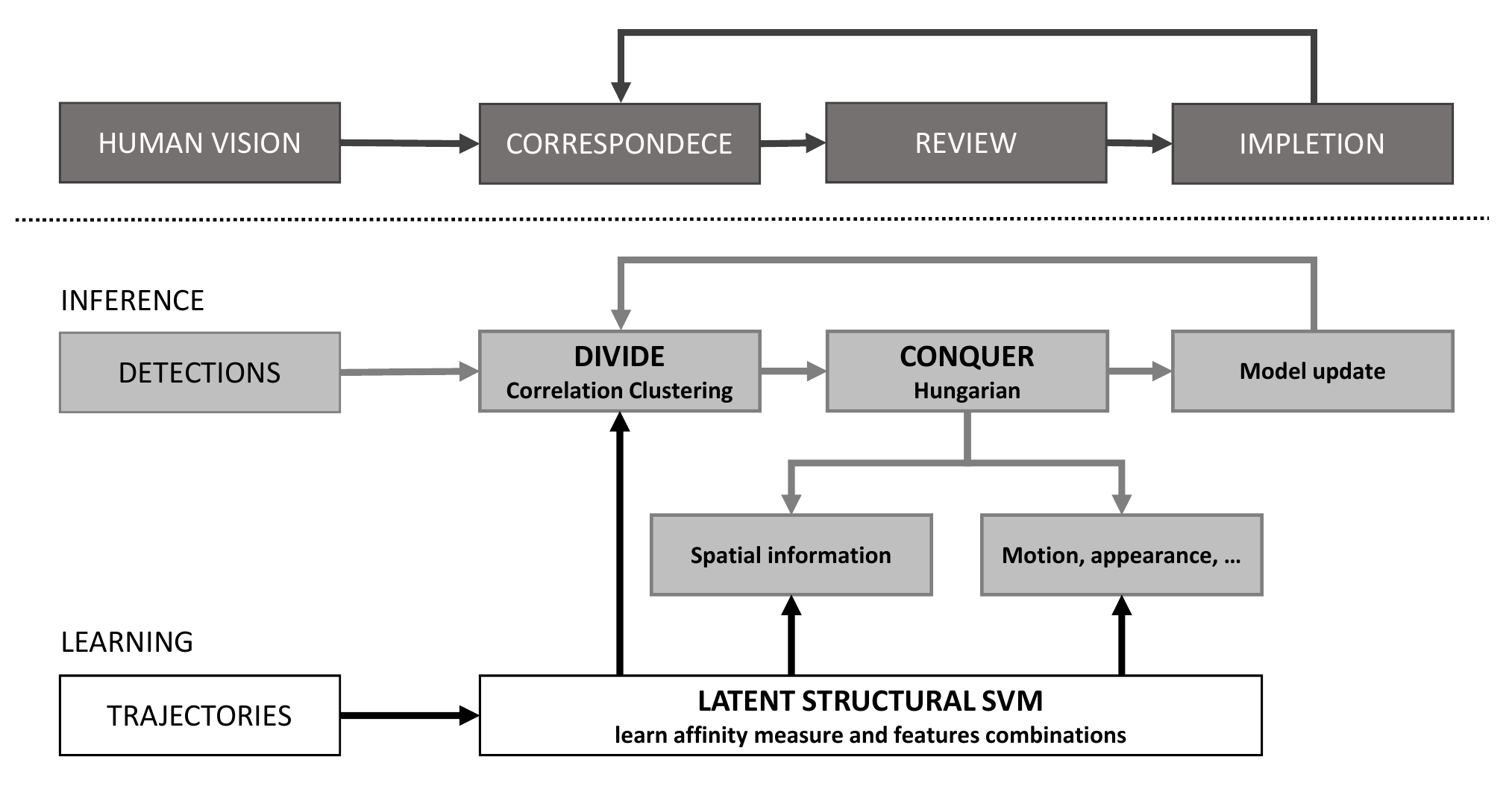}
        \caption{First row shows the human tracking process according to Kahneman, Treisman and Gibbs theory~\cite{kahneman_reviewing_1992}. Below a schematic view of the inference and learning steps underpinning our method.}
        \label{fig:overview}
\end{figure}

The proposed method is inspired by the human cognitive ability to solve the tracking task. In fact, events such as eye movements, blinks and occlusions disrupt the input to our vision system, introducing challenges similar to the ones encountered in real world video sequences and detections.
Perception psychologists have studied the mechanisms employed in our brain during multiple object tracking since the '80s~\cite{kahneman_reviewing_1992,pylyshyn_role_1989,alvarez_how_2007}, though only recently RMI experiments have been used to confirm and validate proposed theories. One of these preeminent theories is given in a 
seminal work by Kahneman, Treisman and Gibbs in 1992~\cite{kahneman_reviewing_1992}.
They proposed the theory of Object Files to understand the dominant role of spatial information in preserving target identity.
The theory highlights the central role of spatial information in a paradigm called Spatio-Temporal Dominance. Accordingly, target correspondence is computed on the basis of spatio-temporal continuity and does not consult non-spatial properties of the target.
If spatio-temporal information is consistent with the interpretation of a continuous target, the correspondence will be established even if appearance features are inconsistent.
%
%
``Up in the sky, look: It's a bird. It's a plane. It's Superman!" - this well known quote, from the respective short animated movie (1941), suggests that the people pointing at Superman changed their visual perception of the target to the extent of giving him a completely different meaning, while they never had any doubt they kept referring to the same object.
Nevertheless, when 
correspondence cannot be firmly established on the basis of spatial information, appearance, motion, and other complex features can be consulted as well.
%
In particular, in \cite{kahneman_reviewing_1992} the tracking process is divided into a circular pipeline of three steps (Fig~\ref{fig:overview}, top row). 
The \emph{correspondence} uses only positional information and aims at establishing if detected objects are either a new target or an existing one appearing at a different location. The \emph{review} activates when ambiguity in assignments arises, and recomputes uncertain target links by also taking into account more complex features.
Eventually, the \emph{impletion} is the final task to assess and induce the perception of targets temporal coherence.

\section{The proposal}
As depicted in Fig.~\ref{fig:overview}, the proposed method relates the 3 steps of \emph{correspondence}, \emph{review} and \emph{impletion} to a \textbf{divide} and \textbf{conquer} approach. 
Targets are divided in the \textit{where} pathway by checking for incongruences in spatial coherence.
Eventually, the tracking solution is conquered by associating coherent elements in the \textit{where} (spatial) domain and incoherent ones in the \textit{what} (visual) domain.

The core of the proposal is twofold.
First, a method to divide potential associations between detections and tracks into local clusters or zones. A zone can be either simple or complex, calling for different features to complete the association. Targets can be directly associated to their closest detections if they are inside a simple zone (\eg when we have the same number of tracks and detections, green area in Fig.~\ref{fig:zones}). Conversely, targets inside complex areas (red in Fig.~\ref{fig:zones}) are subject to a deeper evaluation where appearance, motion and other features may be involved.

Second, we cast the problems of splitting potential associations and solving them by selecting and weighting the features inside a unified structural learning framework that aims at the best set of partitions and adapts from scene to scene.         


\subsection{Problem formulation}
\label{sec:overview}
Online MTT is typically solved by optimizing, at frame $k$, a generic assignment function for a set of tracks $\mathcal{T}$ and current detections $\mathcal{D}_k$:
\begin{equation}
h(\mathcal{T}, \mathcal{D}_k) =\arg\min_{\bf y}\sum_{i=1}^n{\bf C}(i, {\bf y}^i),
\label{eq:assoc}
\end{equation}
where ${\bf y}$ is a permutation vector of $\{1,2,\dots,n\}$ and ${\bf C}\in\mathbb{R}^{n\times n}$ is a cost matrix.
The cost matrix ${\bf C}$ is designed to include dummy rows and columns to account for new detected objects (${\bf D_\text{in}}$) or leaving targets (${\bf T_\text{out}}$). More formally, if matrix ${\bf A} : \mathcal{T}\times \mathcal{D}_k \rightarrow \mathbb{R}$  contains association costs for currently tracked targets and detections, the cost matrix is:
\begin{equation}
{\bf C} = \begin{bmatrix}
{\bf A} & {\bf T_\text{out}}\\
{\bf D_\text{in}} & {\boldsymbol\Xi}
\end{bmatrix}
\label{eq:C}
\end{equation}
where $\bf D_\text{in}$, $\bf T_\text{out}$ contain the cost $\xi$ of creating a new track on the diagonal and $+\infty$ elsewhere. Similarly, $\boldsymbol\Xi$ is a full matrix of value $\xi$.

\begin{figure*}[t]
        \centering
        \begin{subfigure}[b]{0.5\columnwidth}
                \includegraphics[width=\textwidth]{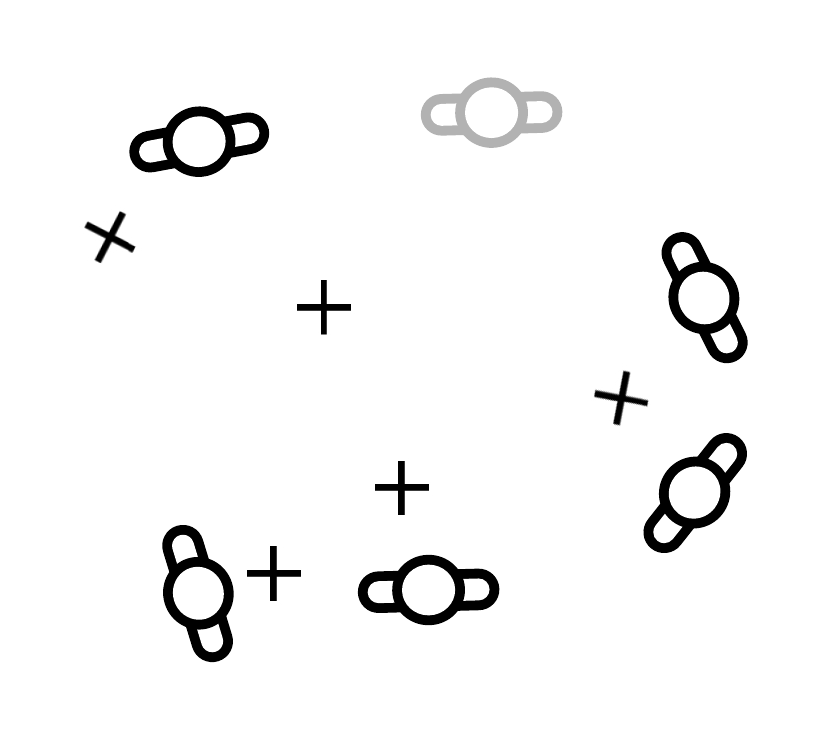}
                \caption{frame input}
        \end{subfigure}\quad\quad\quad\quad\quad
        \begin{subfigure}[b]{0.5\columnwidth}
                \includegraphics[width=\textwidth]{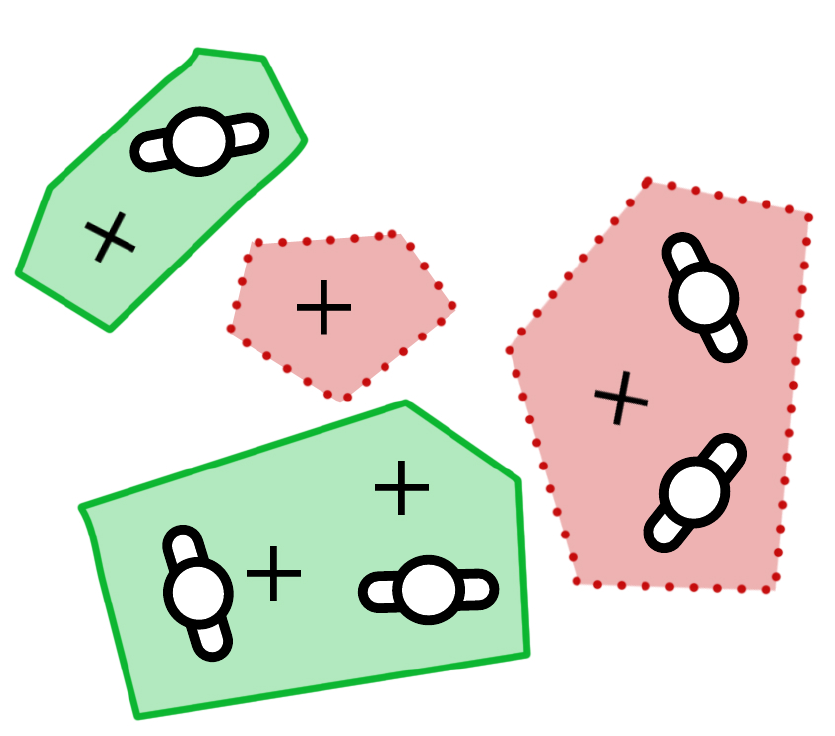}
                \caption{\emph{divide} associations}
               	\label{fig:zones}
        \end{subfigure}\quad\quad\quad\quad\quad
        \begin{subfigure}[b]{0.5\columnwidth}
                \includegraphics[width=\textwidth]{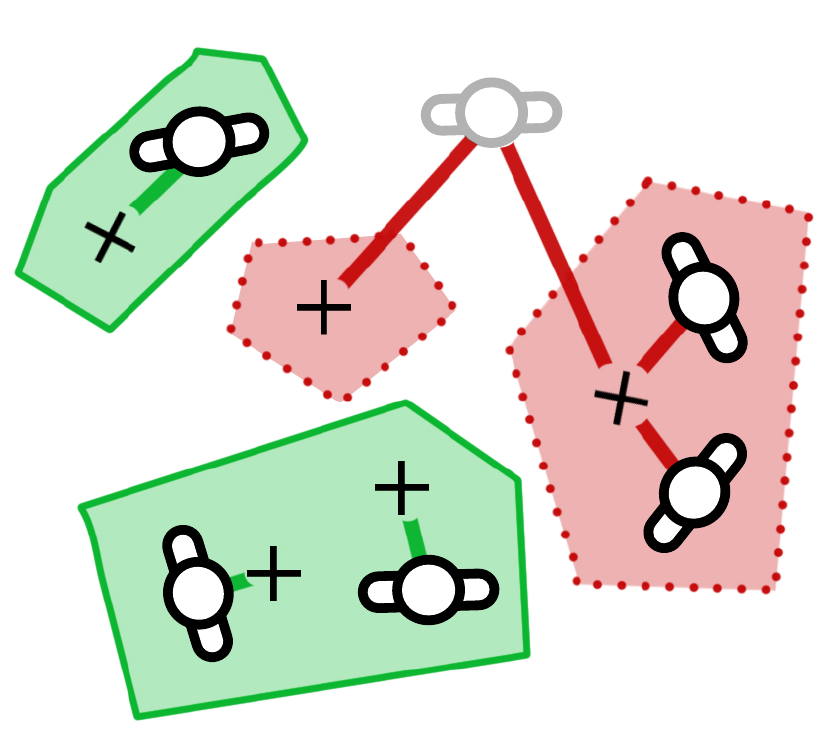}
                \caption{\emph{conquer} associations}
        \end{subfigure}
        \caption{Overview of the inference procedure. (a) In the image targets are represented by bird eye view sketches (shaded when occluded) and detections by crosses. (b) In the \emph{divide} step detections and non-occluded targets are spatially clustered into zones. A zone with an equal number of targets and detections is simple (solid green contours), complex otherwise (dashed red contours). (c) Associations in simple zones are independently solved by means of distance features only. Complex zones are solved by considering more complex features such as appearance or motion and accounting for potentially occluded targets, which are shared across all the complex zones.}
        \label{fig:tracking_results_with_influence_zones}
\end{figure*}

The formulation in Eq.~\eqref{eq:assoc} evaluates all the associations through the same cost function, built upon a preferred set of features.
In order to consider different cost functions for specific subsets of associations, we reformulate Eq.~\eqref{eq:assoc} as:
\begin{equation}
h(\mathcal{T}, \mathcal{D}_k) =\arg\min_{\bf y,\mathcal Z} \sum_{\substack{(i, {\bf y}^i) \in \bf z\\ {\bf z} \in\mathcal{Z}_\text{s}} }{\bf C_s}(i, {\bf y}^i) + \sum_{\substack{ (i, {\bf y}^i) \in \bf z \\ \bf z \in\mathcal{Z}_\text{c}}}{\bf C_c}(i, {\bf y}^i)
\label{eq:assoc_divided}
\end{equation}
where we explicit the different contribution of trivial and difficult associations, whose costs are given by the functions ${\bf C_s}$ and ${\bf C_c}$ respectively. Associations are locally partitioned in zones ${\bf z}\in\mathcal{Z}$ as shown in Fig.~\ref{fig:zones}. Hereinafter, we seamlessly refer to a zone {\bf z} as a portion of the scene or the set of detections and tracks that lie onto it. A zone can be simple ${\bf z} \in \mathcal{Z}_\text{s}$ or complex to solve ${\bf z} \in \mathcal{Z}_\text{c}$ depending on the set of associations it involves.
%

\section{Learning to divide}
\label{sec:divide}
In this section, we propose a method to generate zones ${\bf z}$ and decide whether associations in those zones are simple ${\bf z}\in\mathcal{Z}_s$ or difficult ${\bf z}\in\mathcal{Z}_c$. A zone ${\bf z}$ can be defined as an heterogeneous set of tracks and detections characterized by spatial proximity. Even if simple, the concept of proximity may vary across sequences, and the importance of distances on each axis depends on targets dominant flows in the scene. 
Zones are computed through the Correlation Clustering (CC) method~\cite{bansal_correlation_2002} on the cost matrix ${\bf A}$ suitably modified to obtain an affinity matrix $\bar{\bf A}$ as required by the CC algorithm. To move from cost features (distances) in $\bf A$ to affinity features in $\bar{\bf A}$, the cost features vector is augmented with their similarity counterpart and the affinity value is computed as the scalar product between this vector and a parameter vector ${\boldsymbol\theta}$:
\begin{equation}
\begin{aligned}
&\bar{\bf A}(i,j)= \\
&{\boldsymbol\theta}^T(\underbrace{|t^i_x-d^j_x|, |t^i_y-d^j_y|}_\text{cost features}, \underbrace{1-|t^i_x-d^j_x|, 1-|t^i_y-d^j_y|}_\text{similarity features})^T,
\end{aligned}
\label{eq:param}
\end{equation}
where $t^i$ and $d^j$ are the $i$-th track and $j$-th detection respectively.
The ${\boldsymbol\theta}$ vector has the triple advantage of weighting differently distances on each axis, avoiding to set  thresholds in the affinity computation and controlling the compactness and the balancing of clusters. Further detail on learning ${\boldsymbol\theta}$ are provided in the following sections.

To prevent the creation of clusters composed only of detections or tracks, a symmetric version of $\bar{\bf A}$ is created having a zero block diagonal structure:
\begin{equation}
\bar{\bf A}_\text{sym} = \begin{bmatrix}
{\bf 0} & \bar{\bf A}\\
\bar{\bf A}^T & 0
\end{bmatrix}
\label{eq:Asym}
\end{equation}
Through this shrewdness, two tracks (detections) can be in the same cluster only if close to a common detection (track).
The CC algorithm, applied on $\bar{\bf A}_\text{sym}$, efficiently partition the scene in a set of zones $\mathcal{Z}$ so that the sum of the affinities between track-detection pairs in the same zone is maximized:
\begin{equation}
\arg\max_\mathcal{Z}\sum_{{\bf z}\in\mathcal{Z}}\sum_{(i, j)\in{\bf z}} \bar{\bf A}_\text{sym}(i, j).
\label{eq:CC}
\end{equation}

Eventually, a zone ${\bf z}$ is defined as \textit{simple} if it contains an equal number of targets and detections, otherwise is \textit{complex}. As previously stated, associations in a complex zone ${\bf z}\in\mathcal{Z}_c$ cannot be solved with the use of distance information only (Fig.~\ref{fig:zones}), but require more informative features to disambiguate the decision.

\section{Learning to conquer}
\label{sec:conquer}
The divide mechanism brings the advantage of splitting the problem into smaller local subproblems. Associations belonging to simple zones can be independently solved trough any bipartite matching algorithm. 
The complete tracking problem must deal also with occluded target as well. We consider a target as occluded when it is not associated to a detection (e.g. a miss detection in frame $k$ occurred, shaded people in Fig. \ref{fig:tracking_results_with_influence_zones}).  
Since occluded targets are representation of disappeared objects, they are not included in the zones at the current frame. All the subproblems related to complex zones ${\bf z}\in\mathcal{Z}_c$ are consequently connected by sharing the whole set of occluded targets. 
In order to simultaneously solve the whole set of subproblems, we construct an augmented version of the matrix in Eq.~\eqref{eq:C} where the block ${\bf H}$ accounts for potential associations between occluded tracks and current detections:
\begin{equation}
\widehat{\bf C} = \begin{bmatrix}
\hat{\bf A} & {\boldsymbol +\infty} & {\bf T_\text{out}}\\
{\bf H} & {\bf H_\text{occ}} & {\boldsymbol +\infty}\\
{\bf D_\text{in}} & {\boldsymbol\Xi} & {\boldsymbol\Xi}
\end{bmatrix}.
\label{eq:C_hat}
\end{equation}
${\bf H_\text{occ}}$ is a $\xi$-diagonal matrix ($+\infty$ elsewhere) used to keep occluded tracks still occluded in the current frame. 
The solution of the optimization problem in Eq.~\ref{eq:assoc} on matrix $\widehat{\bf C}$, obtained by applying the Hungarian algorithm, provides the final tracking associations for this frame. 

More precisely, thanks to the peculiar block structure of $\widehat{\bf C}$ a single call to Hungarian results in solving the partitioned association problem in Eq.~\eqref{eq:assoc_divided}, subject to the constraint that each occluded element can be inserted in a single complex zone subproblem solution.
In $\widehat{\bf C}$, simple zones subproblems are isolated
by setting the association cost outside the zone to $+\infty$. Similarly, complex zones results in independent blocks as well, but are connected through the presence of occluded elements, \ie non-infinite entries in ${\bf H}$.\\

\noindent By casting the problem using the cost matrix $\widehat{\bf C}$, it is possible to learn, in a joint framework, to combine features in order to obtain a suitable cost for both the association (either in simple or complex zones) and the partition in zone as well.      
To this end we introduce a linear ${\bf w}$-parametrization on $\hat{\bf A}$ and ${\bf H}$ with a mask vector ${\boldsymbol\pi}_\mathcal{Z}$ that selects the features according to the complexity of the belonging zone :
\begin{equation}
\widehat{\bf C}(i, j) = {\bf w}^T{\boldsymbol\pi}_\mathcal{Z}(i, j)\circ{\bf f}(i, j),
\end{equation}
being $\circ$ the Hadamard product. 
The feature vector contains both simple and complex information between the $i$-th track and the $j$-th detection:
\begin{equation}
\label{eq:feature_vector}
\begin{aligned}
{\bf f}(i, j)^T =~~&(\underbrace{\phantom{||}1\phantom{||}}_{\xi}, \underbrace{|t^i_x-d^j_x|, |t^i_y-d^j_y|}_\text{features for ${\bf z}\in\mathcal{Z}_s$},\\
& \underbrace{1-|t^i_x-d^j_x|, 1-|t^i_y-d^j_y|}_\text{features for \emph{divide} step},\\
&\underbrace{|t^i_x-d^j_x|, |t^i_y-d^j_y|, g_1(t^i, d^j), g_2(t^i, d^j), \dots}_\text{features for ${\bf z}\in\mathcal{Z}_c$}).
\end{aligned}
\end{equation}
where $g_1, g_2, \dots$ are distance functions between track $i$ and detection $j$ on complex features $1$ and $2$ respectively.
Precisely, ${\boldsymbol\pi}_\mathcal{Z}$ selectively activates features according to the following rules:
\begin{equation}
\boldsymbol\pi_\mathcal{Z}(i, j)^T =
\begin{cases}
(~~0, 1, 1, 0, 0, 0, 0, 0, 0, \dots)	&	\text{if (a)}\\
(~~0, 0, 0, 0, 0, 1, 1, 1, 1, \dots)	&	\text{if (b)}\\
(\infty, 0, 0, 0, 0, 0, 0, 0, 0, \dots)	&	\text{if (c)}
\end{cases}
\label{eq:pi_par}
\end{equation}
where the pair target-detection in $\widehat{\bf C}_\mathcal{Z}(i,j)$ may (a) belong to the same simple zone, (b) be composed by elements belonging to complex zones and (c) have elements belonging to different zones.\\
The feature vector ${\bf f}(i, j)$ is computed only on pairs of (possibly occluded) tracks and detections. To extend the parametrization to the whole matrix $\widehat{\bf C}$, it is sufficient to set $\boldsymbol\pi_\mathcal{Z}=(1,0,0,\dots)^T$ outside $\widehat{\bf A}$ and ${\bf H}$. Analogously, for elements $\widehat{\bf C}(i,j)$ outside $\widehat{\bf A}$ or ${\bf H}$, we set ${\bf f}(i,j) = (\infty, 0, 0, \dots)^T$ and ${\bf  f}(i, j) = (1, 0, 0, \dots)^T$ when $\widehat{\bf C}(i, j) = + \infty$ and $\widehat{\bf C}(i,j) = \xi$ respectively.
The learning procedure in Sec. \ref{sec:learning} computes the best weight vector $\bf w$ and consequently $\xi$ is learnt as a bias term. Recall that $\xi$ governs tracks initiation and termination.
Eq.~\eqref{eq:assoc_divided} becomes a linear combination of the weights ${\bf w}$ and a \emph{feature map} $\Phi$:
\begin{equation}
\begin{aligned}
h(\mathcal{T}, \mathcal{D}_k; {\bf w}) &= \arg\max_{{\bf y}, \mathcal{Z}} - {\bf w}^T \sum_{i=1}^n\boldsymbol\pi_\mathcal{Z}(i, {\bf y}^i)\circ{\bf f}(i, {\bf y}^i)\\
&= \arg\max_{{\bf y}, \mathcal{Z}} {\bf w}^T\Phi(\mathcal{T}, \mathcal{D}_k, {\bf y}, \mathcal{Z}).
\end{aligned}
\label{eq:h_par}
\end{equation}
The feature map $\Phi$ is a function evaluating how well the set of zones $\mathcal{Z}$ and the proposed tracking solution ${\bf y}$ for frame $k$ fit on the input data $\mathcal{T}$ and $\mathcal{D}_k$.\\

\noindent Given a set of weights ${\bf w}$, the tracking problem in Eq.~\eqref{eq:h_par} can be solved by first computing the zones $\mathcal{Z}$ through the \emph{divide} step on matrix $\bar{\bf A}_\text{sym}$ of Eq.~\eqref{eq:Asym} and then by \emph{conquering} the associations in each zone through the Hungarian method on matrix $\widehat{\bf C}$. Note that now $\bar{\bf A}_\text{sym}(i,j)={\bf w}^T (0,1,1,1,1,0,0,\dots)^T\circ{\bf f}(i,j)$ and ${\boldsymbol\theta}$ is a subset of $\bf w$.

\begin{algorithm*}[t!]
\caption{Block-Coordinate Primal-Dual Frank-Wolfe Algorithm for learning ${\bf w}$ on a sequence of $K$ frames}
\label{alg1}                           
\begin{algorithmic}[1]                    
    \STATE Let ${\bf w}^{(0)} \leftarrow {\bf 0}, {\bf w}_k^{(0)} \leftarrow {\bf 0}, l^{(0)} \leftarrow 0, l_k^{(0)} \leftarrow 0$ for $k=1,\dots,K$
	\FOR{$k \leftarrow 1$ \TO $K$ }
		\STATE {Compute simple features for learning to \emph{divide} Eq. \eqref{eq:feature_vector}}
		\STATE {\bf Latent completion}: $\mathcal{Z}_k=\arg\max_{\mathcal{Z}}{\bf w}^T\Phi(\mathcal{T}, \mathcal{D}_k, {\bf y}_k, \mathcal{Z})$ through \emph{Correlation Clustering} on $\bar{\bf A}_\text{sym}$ of Eq.~\eqref{eq:Asym}
		\STATE {Compute complex features for learning to \emph{conquer} Eq. \eqref{eq:feature_vector}}
		\STATE {\bf Max Oracle}: ($\bar{\bf y}_k, \bar{\mathcal{Z}}_k) = \arg\max_{{\bf y},\mathcal{Z}} H_k({\bf y}, \mathcal{Z}; {\bf w})$ through \emph{Hungarian} on Eq.~\eqref{eq:oracle_loss}
		\STATE Let ${\bf w}_s \leftarrow \frac{1}{\lambda K}\psi_k(\bar{\bf y}, \bar{\mathcal{Z}})$ and $l_s \leftarrow \frac{1}{n}\Delta_k(\bar{\bf y}, \bar{\mathcal{Z}})$
		\STATE Let $\gamma \leftarrow [\lambda({\bf w}_k^{(r)}-{\bf w}_s)^T{\bf w}^{(r)}-l_k^{(r)}+l_s]/[\lambda\|{\bf w}_k^{(r)}-{\bf w}_s\|^2]$ and clip to $[0,1]$
		\STATE Update ${\bf w}_k^{(r+1)} \Leftarrow (1-\gamma){\bf w}_k^{(r)} + \gamma {\bf w}_s$ and $l_k^{(r+1)}\Leftarrow (1-\gamma)l_k^{(r)}+\gamma l_s$
		\STATE Update ${\bf w}^{(r+1)}\Leftarrow {\bf w}^{(r)} + {\bf w}_k^{(r+1)} - {\bf w}_k^{(r)}$ and $l^{(r+1)} = l^{(r)} + l_k^{(r+1)}-l_k^{(r)}$
	%
	\ENDFOR
\end{algorithmic}
\end{algorithm*}

\section{Online subgradient optimization}
\label{sec:learning}
The problem of Eq.~\eqref{eq:h_par} requires to identify the complex structured object $({\bf y}, \mathcal{Z})\in\mathcal{Y}\times\mathcal{\bf Z}$ such that $\mathcal{Z}$ is the set of zones that best explain the $k$-th frame tracking solution ${\bf y}$ for an input $(\mathcal{T}, \mathcal{D}_k)$. Zones ${\bf z}\in\mathcal{Z}$ are modelled as latent variables, since they remain unobserved during training. To this end, we learn the weight vector ${\bf w}$ in $h(\mathcal{T}, \mathcal{D}_k; {\bf w})$ through Latent Structural SVM~\cite{yu_learning_2009} by solving the following unconstrained optimization problem over the training set $\mathcal{S}=\{(\mathcal{T}, \mathcal{D}_k, {\bf y}_k)\}_{k=1\dots K}$:
\begin{equation}
\label{eq:lssvm}
\min_{\bf w} \frac{\lambda}{2}\|{\bf w}\|^2 + \frac{1}{K}\sum_{k=1}^{K}\tilde{H}_k({\bf w}),
\end{equation}
with $\tilde{H}_k({\bf w})$ being the \emph{structured hinge-loss}.
$\tilde{H}_k({\bf w})$ results from solving the loss-augmented maximization problem
\begin{equation}
\label{eq:oracle}
\begin{aligned}
\tilde{H}_k({\bf w}) &= \max_{{\bf y},\mathcal{Z}} H_k({\bf y}, \mathcal{Z}; {\bf w})\\
& = \max_{{\bf y},\mathcal{Z}} \Delta_k({\bf y}, \mathcal{Z}) - \langle {\bf w}, \psi_k({\bf y}, \mathcal{Z})\rangle,
\end{aligned}
\end{equation}
where $\Delta_k({\bf y}, \mathcal{Z})=\Delta({\bf y}_k, \mathcal{Z}_k, {\bf y}, \mathcal{Z})$ is a loss function that measures the error of predicting the output ${\bf y}$ instead of the correct output ${\bf y}_k$ while assuming $\mathcal{Z}$ to hold instead of $\mathcal{Z}_k$, and we defined $\psi_k({\bf y}, \mathcal{Z}) = \Phi(\mathcal{T}, \mathcal{D}_k, {\bf y}_k, \mathcal{Z}_k) - \Phi(\mathcal{T}, \mathcal{D}_k, {\bf y}, \mathcal{Z})$ for notation convenience.

Solving Eq.~\eqref{eq:oracle} is equivalent to finding the output-latent pair $({\bf y}, \mathcal{Z})$ generating the most violated constraint, for a given input $(\mathcal{T}, \mathcal{D}_k)$ and a latent setting $\mathcal{Z}_k$. 
\begin{figure}[t]
\centering
        \begin{subfigure}[b]{0.23\textwidth}
                \includegraphics[width=\textwidth]{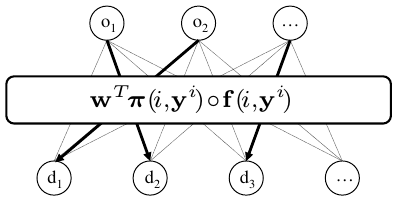}
                \caption{Inference Step}
                \label{fig:inference_oracle_inference}
        \end{subfigure}
        ~
        \begin{subfigure}[b]{0.23\textwidth}
                \includegraphics[width=\textwidth]{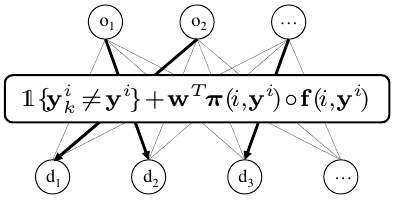}
                \caption{Maximization Oracle}
        \label{fig:inference_oracle_oracle}
        \end{subfigure}
        \caption{Thanks to the choice of the Hamming loss, the maximization oracle is reduced to an assignment problem efficiently solved through the Hungarian algorithm, as for the inference step. \vspace{-0.3cm}}
\end{figure}
Despite the generality of the learning framework, the loss function $\Delta$ is problem dependent and must be accurately chosen. In particular, we adopted the Hamming loss function that, substituted in Eq.~\eqref{eq:oracle}, behaves linearly making the maximization oracle solvable as a standard assignment problem, Fig.~\ref{fig:inference_oracle_oracle}:
\begin{equation}
\label{eq:oracle_loss}
\tilde{H}_k({\bf w}) = \max_{{\bf y},\mathcal{Z}} \sum_{i=1}^n\mathds{1}\{{\bf y}_k^i\neq{\bf y}^i\} + {\bf w}^T\boldsymbol\pi_\mathcal{Z}(i, {\bf y}^i)\circ{\bf f}(i, {\bf y}^i)
\end{equation}
where $\Phi(\mathcal{T}, \mathcal{D}_k, {\bf y}_k, \mathcal{Z}_k)$ was dropped as not dependent on either ${\bf y}$ or $\mathcal{Z}$.\\
\noindent The learning step of Eq.~\eqref{eq:lssvm} can be efficiently solved online, under the premise that the dual formulation of LSSVM results in a continuously differentiable convex objective after latent completion.
We designed a modified version of the Block-Coordinate Frank-Wolfe algorithm~\cite{lacoste-julien_block-coordinate_2013} presented in Alg.~\ref{alg1}.
The main insight here is to notice that the linear subproblem employed by Frank-Wolfe (line 5) is equivalent to the loss-augmented decoding subproblem of Eq.~\eqref{eq:oracle_loss}, which can be solved efficiently through the Hungarian algorithm~\cite{kuhn_hungarian_1955}.
To deal with latent variables during optimization, we added the latent completion process (line 4) where, given an input/output pair, the latent variable $\mathcal{Z}_k$ which best explain the solution ${\bf y}_k$ to the observed data is found.
Through the latent completion step, the objective function optimized by Frank-Wolfe has guarantees to be convex.

\begin{figure*}[t!]
        \centering
        \begin{subfigure}[b]{89px}
                \includegraphics[height=50px]{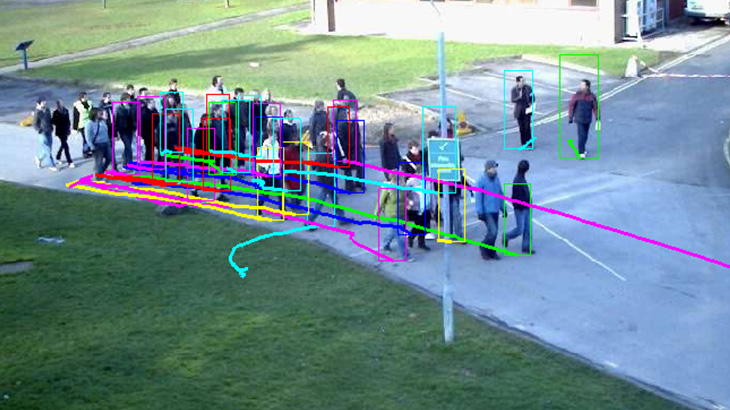}
        \end{subfigure}~
        \begin{subfigure}[b]{58px}
                \includegraphics[height=50px]{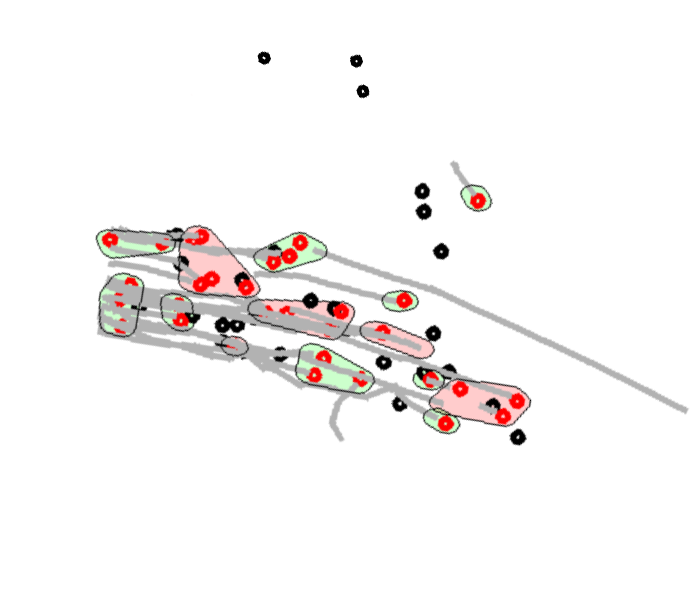}
        \end{subfigure}~~~~~~~
        \begin{subfigure}[b]{89px}
                \includegraphics[height=50px]{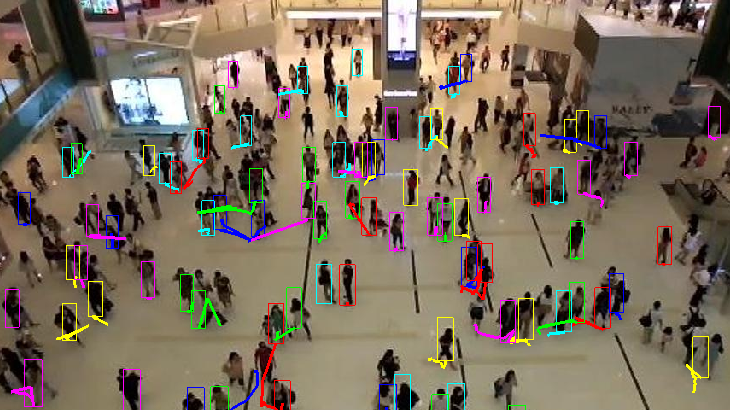}
        \end{subfigure}~
        \begin{subfigure}[b]{58px}
                \includegraphics[height=50px]{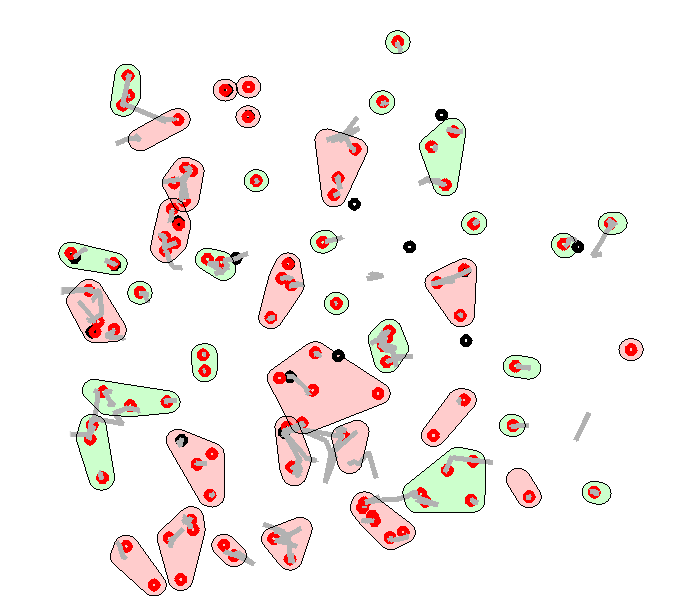}
        \end{subfigure}~~~~~~~
        \begin{subfigure}[b]{89px}
                \includegraphics[height=50px]{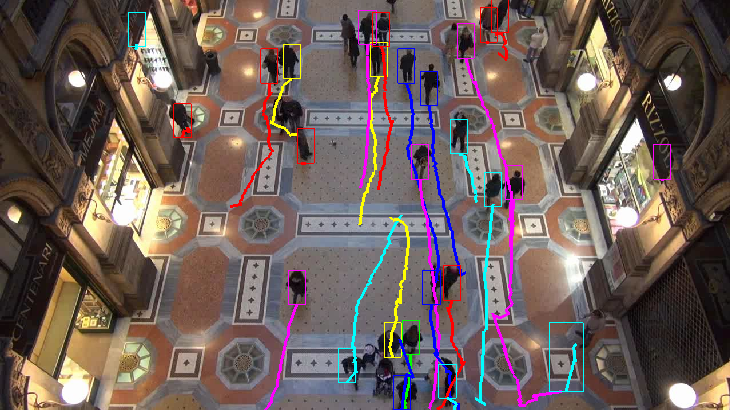}
        \end{subfigure}~
        \begin{subfigure}[b]{58px}
                \includegraphics[height=50px]{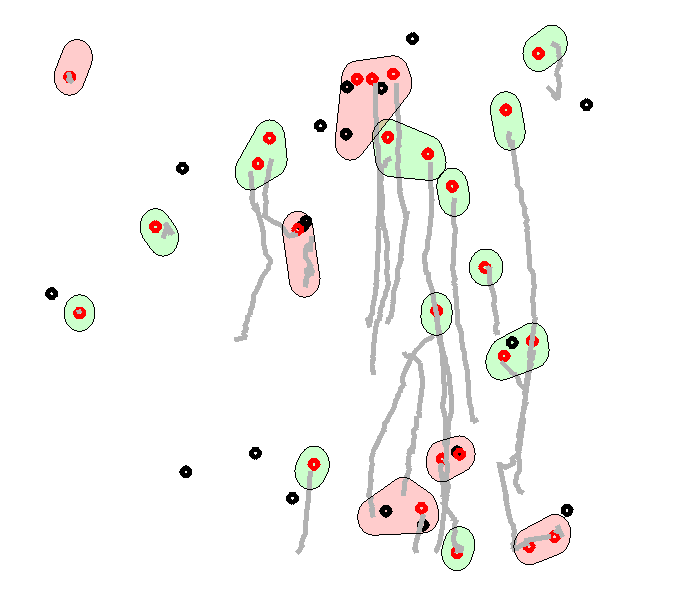}
        \end{subfigure}\\
        \begin{subfigure}[b]{89px}
                \includegraphics[height=50px]{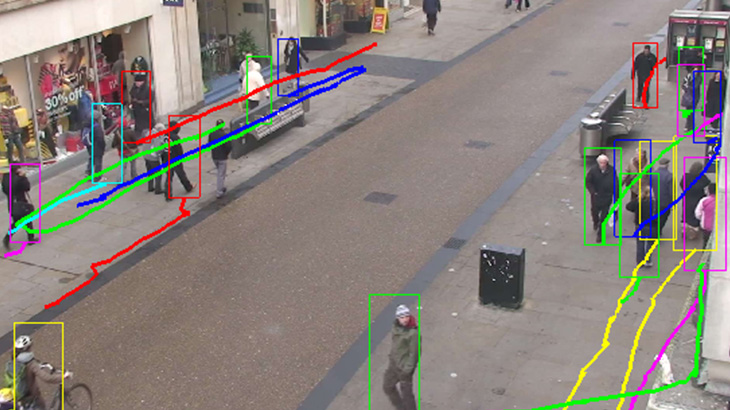}
        \end{subfigure}~
        \begin{subfigure}[b]{58px}
                \includegraphics[height=50px]{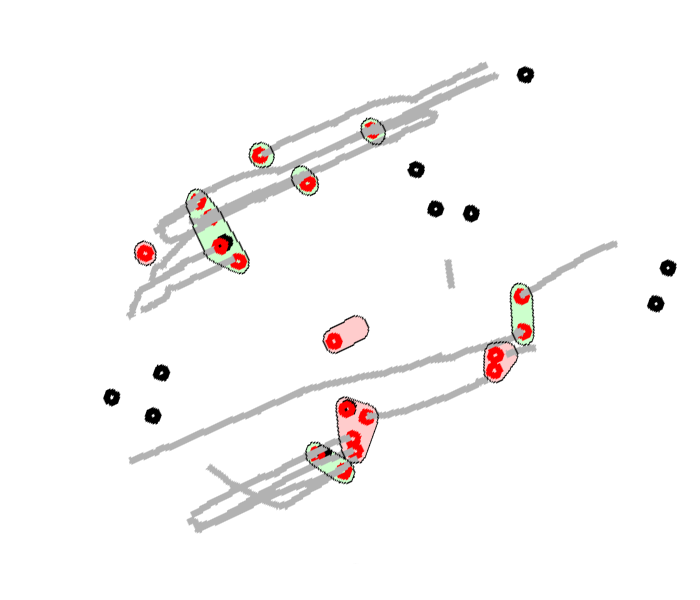}
        \end{subfigure}~~~~~~~~~~
        \begin{subfigure}[b]{89px}
                \includegraphics[height=50px]{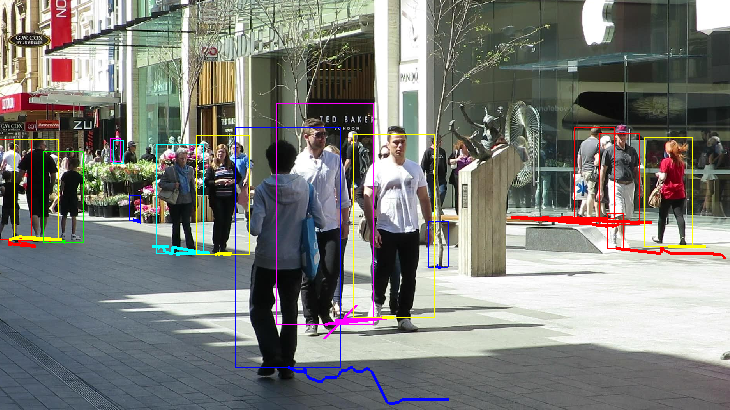}
        \end{subfigure}~
        \begin{subfigure}[b]{58px}
                \includegraphics[height=50px]{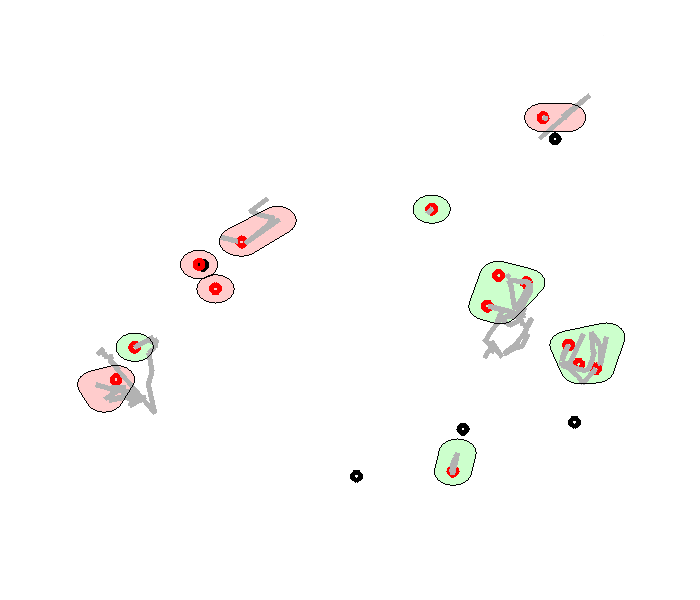}
        \end{subfigure}~~~~~~~~~~
        \begin{subfigure}[b]{89px}
                \includegraphics[height=50px]{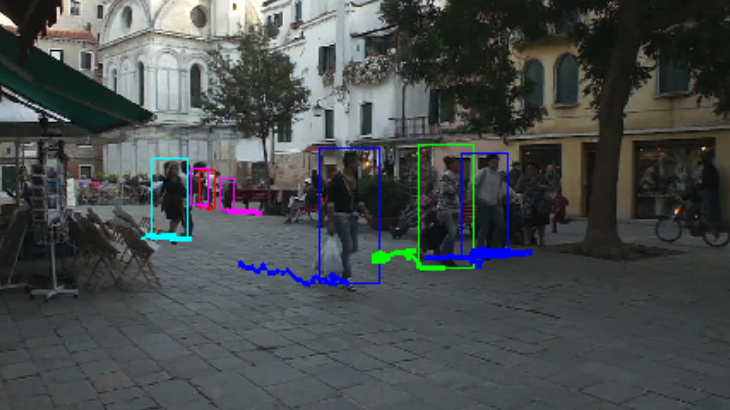}
        \end{subfigure}~
        \begin{subfigure}[b]{58px}
                \includegraphics[height=50px]{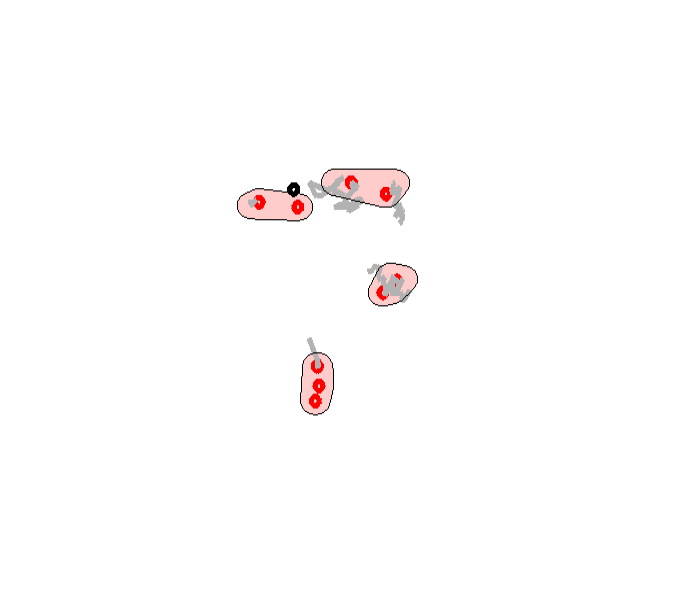}
        \end{subfigure}~~~~~~~
        \caption{Tracking results on PETS09-S2L3, 1shatian3 and GVEII from the MCD dataset (top row). AVG-TownCentre, ADL-Rundle-3 and Venice-1 from the MOT Challenge sequences (bottom). Next to images, simple (green) and complex (red) zones are displayed.}
        \label{fig:visual_tracking}
\end{figure*}

\section{Experimental results}
In this section we present two different experiments that highlight the improvement of our method over state of the art trackers in static camera sequences.
The first experiment is devoted to stress the method in clutter scenarios where moderate crowd occurs and our divide and conquer approach gives its major benefits in terms of both computational speed and performances. The second experiment is on the publicly available \emph{MOT Challenge} dataset that is becoming a standard for tracking by detection comparison.
Test were evaluated employing the CLEAR MOT~\cite{bernardin2008evaluating} measures and trajectory based measures (MT,ML,FRG) as suggested in~\cite{motchallenge}. 
All the detections, where not provided by authors, have been computed using the method in \cite{dollar_fast_2014} as suggested by the protocol in \cite{motchallenge}.
Results are averaged per experiment in order to have a quick glimpse on the tracker performances. Individual sequences results are provided in the additional material.
To train the parameters acting on the complex zones, the LSSVM have been trained with ground truth (GT) trajectories and the addition of different levels of random noise simulating miss and false detections.
In all the tests, occluded objects locations are updated in time using a Kalman Filter with a constant velocity state transition model, and discarded if not reassociated after 15 frames.

\subsection{Features}
The strength of the proposal is the joint LSSVM framework that learns to weight features for both partitioning the scene and associating targets. On these premises, we purposely adopted standard features. Without loss of generality, the method can be expanded through additional and more complex features as well. The features always refer to a single detection ${\bf d}\in\mathcal{D}_k$ and a single track ${\bf t}\in\mathcal{T}$, occluded or not, and its associated history, in compliance with Eq.~\eqref{eq:feature_vector}.

In the experiments, the appearance of the targets is modeled through a color histogram in the RGB space.
Every time a new detection is associated to a track, its appearance information is stored in the track history. The appearance feature $g_1$ is then computed as the average value of the Kullback-Leibler distance of the detection histogram from track previous instances.
Additionally, we designed tracks to contain their full trajectories over time. By disposing of the trajectories, we modeled the motion coherence $g_2$ of a detection w.r.t a track by evaluating the smoothness of the manifold fitted on the joint set of the new detected point and the track spatial history. More precisely, given a detected point, an approximate value of the Ricci curvature is computed by considering only the subset of detections of the trajectory lying inside a given neighborhood of the detected point. An extensive presentation of this feature is in~\cite{gong2012}.

\begin{figure*}[t!]
        \begin{subfigure}[c]{0.65\columnwidth}
                \includegraphics[height=4.7cm]{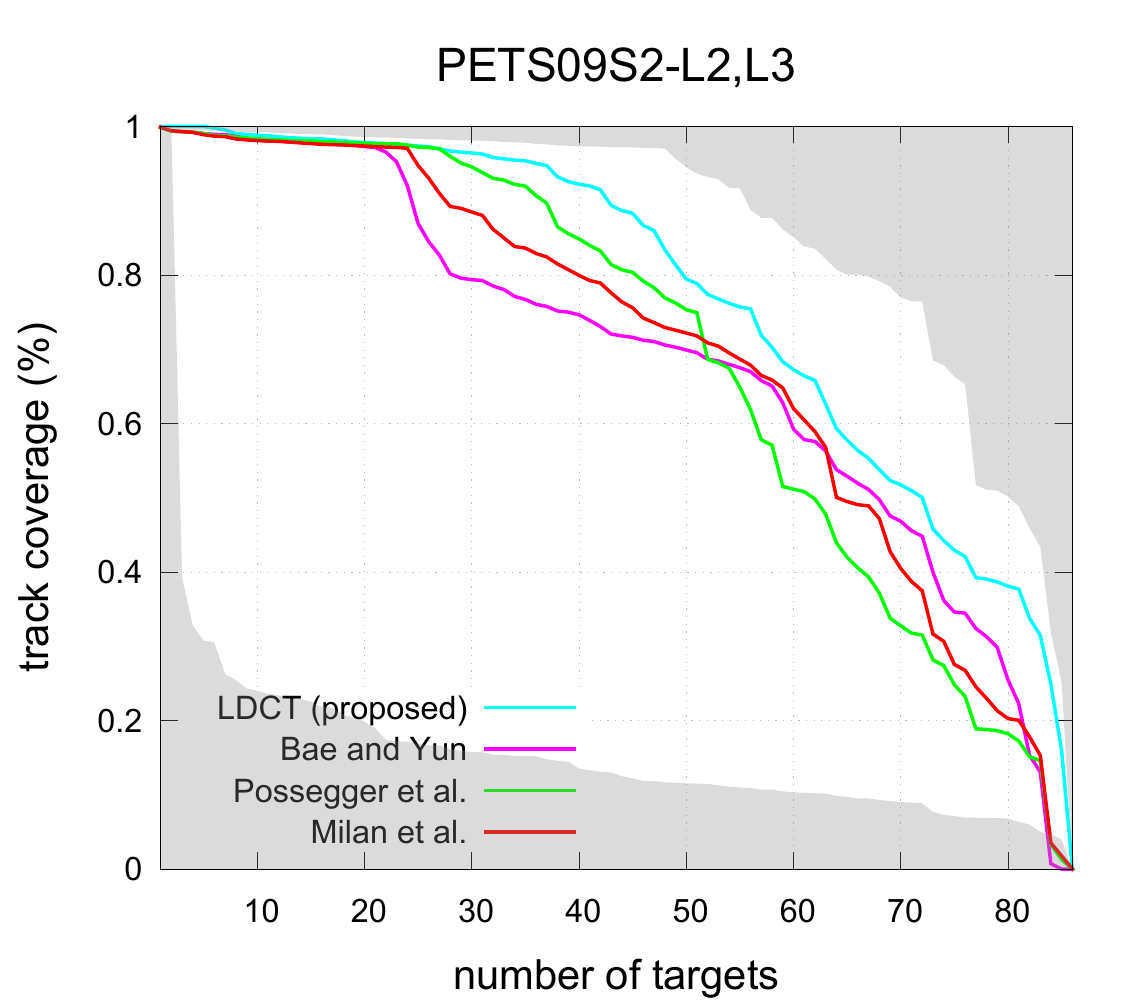}
        \end{subfigure}~~~~~
        \begin{subfigure}[c]{0.65\columnwidth}
                \includegraphics[height=4.7cm]{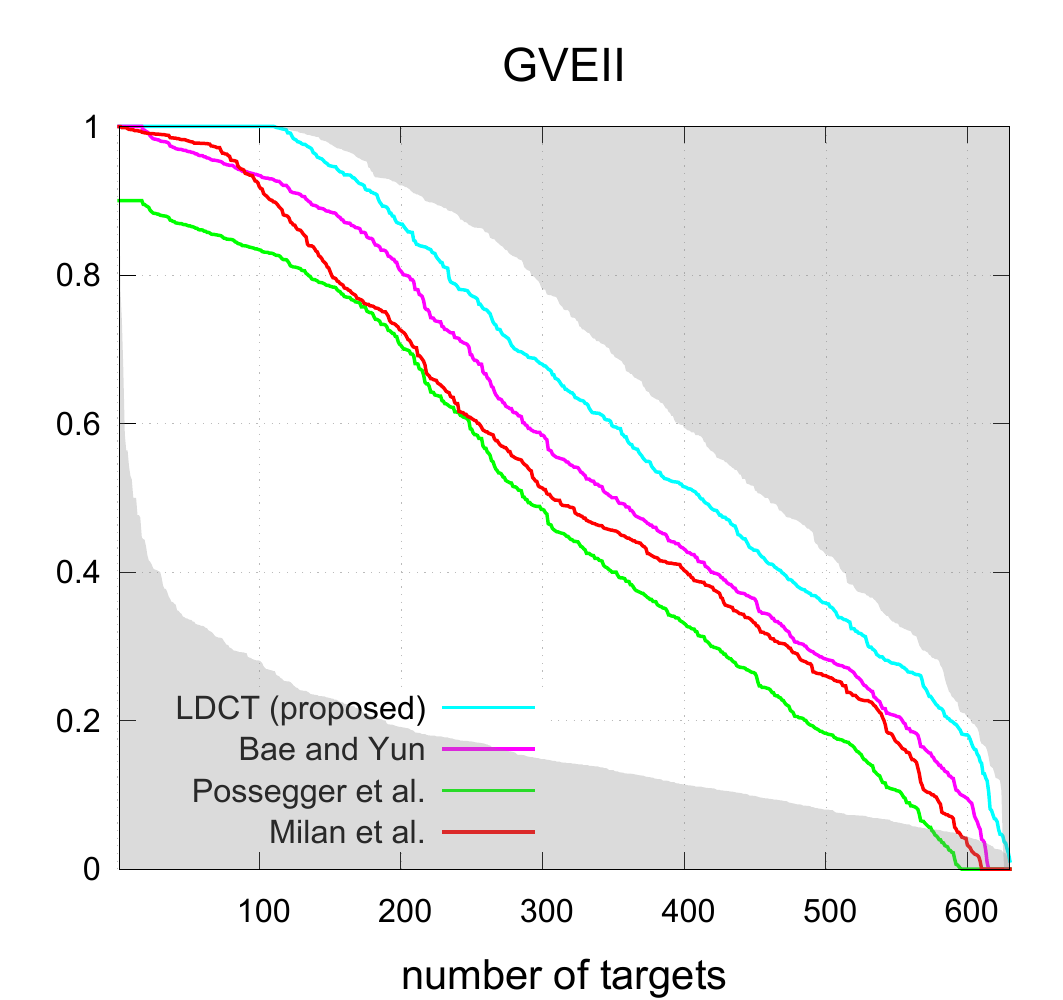}
        \end{subfigure}~
        \begin{subfigure}[c]{0.65\columnwidth}
                \includegraphics[height=4.7cm]{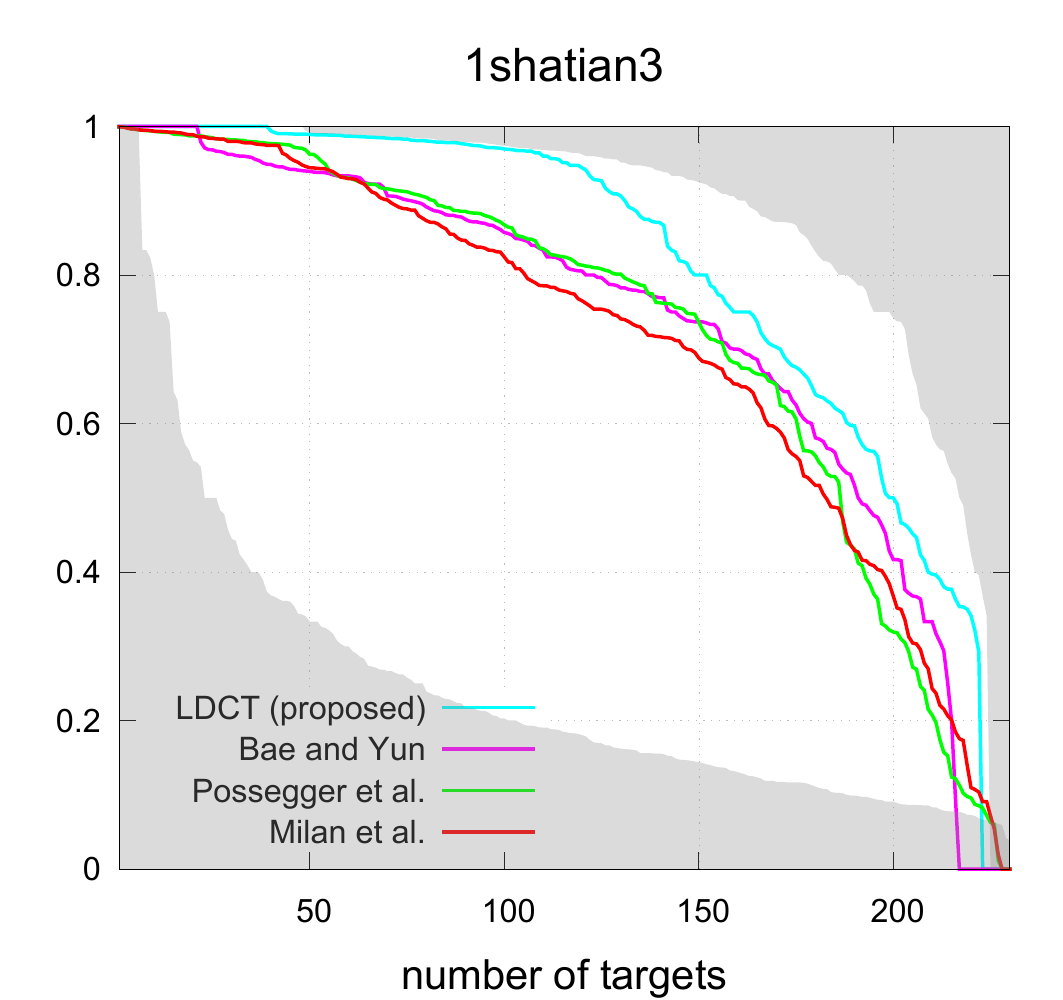}
        \end{subfigure}
        \caption{Tracks length curves (TL) on MCD sequences. The gray shaded area indicates the performances reached by a simple global NN algorithm (lower bound) and the highest score obtained for each track combining all different methods results (upper bound).}\label{fig:survival_curves}
\end{figure*}

\subsection{Datasets and Settings}
\noindent {\bf Midly Crowded Dataset (MCD)}: the dataset is a collection of moderately crowded videos taken from both public benchmarks with the addition of ad-hoc sequences. This dataset consists of 4 sequences: the well-known PETS09 S2L2 and S2L3 sequences, and 2 new sequences. GVEII is characterized by a high number of pedestrian crossing the scene (up to 107 people per frame), while 1shatian3, captured by~\cite{6714561}, is a sequence characterized by a high density and clutter (up to 227 people per frame). A single training stage was performed by gathering the first $30\%$ of each video. These frames have not been used at test time.\\

\noindent {\bf MOT Challenge}: the dataset consists of several public available sequences in different scenarios. Detections and annotations are provided by the MOTChallenge website. In our test we consider the subset of the sequences coming from fixed cameras since distances are not meaningful in the moving camera settings: TUD-Crossing, PETS09-S2L2, AVG-TownCentre, ADL-Rundle-3, KITTI-16 and {Venice-1}.\\Learning was performed on a distinct set of sequences provided on the website for training.

\begin{table}
\scalebox{0.75}{
\begin{tabular}{|l|c|c|c|c|c|c|c|}
\hline
 &  app & MOTA & MOTP & MT & ML & IDS & FRG \\
\hline
\rowcolor{Gray}
{\bf LDCT}  & \emph{w.n.} & {\bf 47.7} & {\bf 68.8} & {\bf 88} & {\bf 26} & {\bf 209} & {\bf 103}\\
\hline
\rowcolor{Gray2}
{\bf LDCT} (all features)  & \checkmark & 40.6 & 66.3 & 61 & 43 & 446 & 193\\
\hline
\rowcolor{Gray2}
 {\bf LDCT} (only simple) &  & 36.4 & 64.7 & 58 & 50 & 586 & 276\\
\hline
Bae and Yun~\cite{bae_robust_2014}   & \checkmark & 39.0 & 65.8 & 84 & 35 & 637 & 289  \\ 
\hline
Possegger \emph{et al.}~\cite{possegger_occlusion_2014}  &  & 38.7 & 65.0 & 79 & 37 & 455 & 440  \\ 
\hline
Milan \emph{et al.}~\cite{milan_continuous_2014} & & 40.6 & 66.7 & 64 & 42 & 242 & 141 \\ 
\hline
\end{tabular}
}
\caption{Average results on MCD. In the appearance column, \emph{w.n.} is \emph{when needed}. More details on the light gray baselines in the text.}
\label{tab:results}
\end{table}

\begin{table}
\scalebox{0.75}{
\begin{tabular}{|l|c|c|c|c|c|c|c|c|}
\hline
 & MOTA & MOTP & MT & ML & FP & FN & IDS & FRG \\
\hline
\multicolumn{9}{|l|}{\emph{Online}}\\
\hline
\hline
\rowcolor{Gray2}
{\bf LDCT} & {\bf 43.1} & {\bf 74.5} & {\bf 9} & {\bf 10} & 682 & {\bf 2780} & 161 & 187 \\
\hline
RMOT & 30.4 & 70.2 & 2 & 27 & 1011 & 3259 & 74 & 125 \\ 
\hline
TC\_ODAL & 24.2 & 70.9 & 1 & 31 & 1047 & 3528 & 75 & 152 \\ 
\hline
\multicolumn{9}{|l|}{\emph{Offline}}\\
\hline
\hline
MotiCon & 32.0 & 70.6 & 2 & 30 & 777 & 3280 & 110 & 105 \\ 
\hline
SegTrack & 32.3 & 72.1 & 3 & 38 & 520 & 3454 & 80 & 76 \\ 
\hline
CEM & 28.1 & 71.2 & 5 & 24 & 1256 & 3088 & 87 & 97 \\ 
\hline
SMOT & 23.9 & 71.7 & 2 & 27 & 706 & 3627 & 120 & 208 \\ 
\hline 
TBD & 28.0 & 71.3 & 3 & 25 & 1233 & 3083 & 192 & 193 \\ 
\hline 
DP\_NMS & 22.7 & 71.4 & 3 & 17 & 1062 & 3052 & 529 & 325 \\ 
\hline 
\end{tabular} 
}
\caption{Averaged results of our method (LDCT) and the other MOT Challenge competitors on the 6 fixed camera sequences. See: \url{http://www.motchallenge.net} for detailed results.}
\label{tab:motchallenge}
\end{table}

\subsection{Comparative evaluation}
\noindent {\bf Results on MCD}: Quantitative results of our proposal on the MCD dataset compared with the state of the art trackers are presented in Tab.~\ref{tab:results}, while visual results are in Fig.~\ref{fig:visual_tracking}.
We compared against two very recent online methods \cite{possegger_occlusion_2014,bae_robust_2014} that focus either on target motion or appearance. Moreover, the offline method~\cite{milan_continuous_2014} has been considered being one of the most effective MTT methods up to now.
In the MCD challenging sequences, we outperform the competitors in terms of MT values having also the lowest number of IDS and FRAG. This is basically due to the selective use of the proper features depending on the outcomes of the divide phase of our algorithm. This solution allows our tracker to take the best of both worlds against \cite{possegger_occlusion_2014} and \cite{bae_robust_2014}. MOTA measure is higher as well testifying the overall quality of the proposed tracking scheme.
Additionally, in Fig.~\ref{fig:survival_curves} we reported the track lenght curves (TL) on the MCD dataset. TL curve is computed by considering the length of the correctly tracked GT trajectories plotted in descending order. The plot gives information on the ability of the tracker to build continuous and correct tracks for all the ground truth elements in the scene, neglecting the amount of false tracks inserted. Our AUC is always greater than competitors' thanks to the adoption of complex zones that effectively deals with occluded/disapperared objects and keep the tracks longer.

To evaluate the improvement due to the adoption of the divide and conquer steps, which is the foundation of our tracker, in Tab.~\ref{tab:results} we also test two baselines: when either all features or spatial features only were used for all the assignments independently of the zone type. In both tests, the divide step, the parameter learning and occlusion handling remain as previously described.
Improvement of the complete method (dark gray) over these baselines (light gray) suggests that complex features are indeed more beneficial when used selectively.
\\

\noindent {\bf Results on MOT Challenge}:
Tab \ref{tab:motchallenge} summarizes the accuracy of our method compared to other state of the art algorithms on the MOT Challenge dataset. Similarly to the MCD experiment, we observe that our algorithm outperforms the other state of the art methods.
Our method achieves best results in most of the metrics, keeping IDS and FRG relatively low as well. In turn, our method records the highest MOTA compared to others with a significant margin (+10\%). Excellent results on this dataset highlight the generalization ability of our method, which was trained on sequences different (although similar) from the ones in the test evaluation. Fig.~\ref{fig:visual_tracking} shows some qualitative examples of our results.
\\
Furthermore, our online tracker has been designed to perform considerably fast. We report an average performances of 10~fps on the MOT Challenge sequences. The runtime is strongly influenced by the number of detections as well as by the number of tracks created up to a specific frame. The performances are in line or faster than the majority of the current methods that report an average of 3-5~fps.

The computational complexity of solving Eq.~\eqref{eq:assoc} using the Hungarian algorithm is $\mathcal{O}(N+N_o)^3$ with $N$ the number of tracks and detections to be associated and $N_o$ the number of occluded tracks.
Since the complexity of the divide step is linear in the number of targets, our algorithm reduced the assignment complexity to $N\mathcal{O}(\frac{\alpha}{2})+ \mathcal{O}(N\beta+N_o)^3$. The first term applies for simple zones and is linear in $N$ being dominated by $\alpha$ that is the average number of detections in every partition ($\alpha<<N$). The second term modulates the complexity of the association algorithm in complex zones by the $\beta$ factor, \ie is the percentage of complex zones in the scene. Eventually the $N_o$ term is related to the recall of the chosen detector. As an example $N_o$ can be realistically set to $0.3N$ and, if the percentage of complex zones $\beta$ is $10\%$, the algorithm is $50\times$ faster than its original counterpart.

\section{Conclusion}
In this work, we proposed an enhanced version of the Hungarian online association model to match recent features advancement and cope with different sequences peculiarities. 
The algorithm is able to learn to effectively partition the scene and choose the proper feature combination to solve simple and complex association in an online fashion.  
As observed in the experiments, the benefits of our divide and conquer approach are evident in terms of both computational complexity of the problem and tracking accuracy.  

The proposed tracking framework can be extended/enriched with a different set of simple and complex features and it can learn to identify the relevant ones for the specific scenario\footnote{Although analogy with cognitive theory holds for spatial features only.}. This can open a major room for improvement by allowing the community to test the method with more complex and sophisticated features. We invite the reader to download the code and to test it by adding her favorite features.
\newpage

\section*{Appendix A: Block-Coordinate Frank-Wolfe optimization of Latent Structural SVM}

In a recent paper by Lacoste-Julien \emph{et al.}~\cite{lacoste-julien_block-coordinate_2013} the efficient use of Block-Coordinate Frank-Wolfe optimization for the training of structural SVM was demonstrated. They noted that by disposing of a maximization oracle, subgradient methods could be applied to solve the non-smooth unconstrained problem of Eq.~\eqref{eq:ssvm}. The notation follows the one used in the paper.
\begin{equation}
\label{eq:ssvm}
\min_{\bf w}\frac{\lambda}{2}\|{\bf w}\|^2 + \frac{1}{K}\sum_{k=1}^K\tilde{H}_k({\bf w}),
\end{equation}
where $\tilde{H}_k({\bf w})$ is exactly the optimal value for the necessary max oracle. The Lagrange dual of the above $K$-slack formulation of Eq.\eqref{eq:ssvm} has $m=\sum_k|\mathcal{Y}_k|$ potential support vectors. Writing $\alpha_k({\bf y})$ for the dual variables associated with the the $k$-th training example and potential output ${\bf y}$, the dual problem is given by
\begin{equation}
\label{eq:dualssvm}
\begin{aligned}
\min_{{\boldsymbol \alpha}\in[0,1]^m}&f(\boldsymbol\alpha) = \frac{\lambda}{2}\|A\boldsymbol\alpha\|^2 - {\bf b}^T\boldsymbol\alpha\\
\text{s.t.}&\sum_{{\bf y}\in\mathcal{Y}_k}\alpha_k({\bf y})=1,\forall k\in\mathbb{N}_K
\end{aligned}
\end{equation}
Here matrix $A$ and vector ${\bf b}$ are constructed by simple Lagrangian derivation.
The only two requirements that need to be satisfied in order to apply Frank-Wolfe algorithm on the problem of Eq.\eqref{eq:dualssvm} are:
 \begin{itemize}
 \item the domain $\mathcal{M}$ of $f(\boldsymbol\alpha)$ has to be compact, and
 \item the convex objective $f$ has to be continuously differentiable.
\end{itemize}
Observe that the domain $\mathcal{M}$ of Eq.~\eqref{eq:dualssvm} is the product of $K$ probability simplices, $\mathcal{M}=\Delta_{|\mathcal{Y}_1|}\times\dots\times\Delta_{|\mathcal{Y}_K|}$. It is thus compact by the geometrical definition of simplex. We now present matrix $A$ and vector ${\bf b}$ for the latent formulation of SSVM and check that $f$ is continuously differentiable. Recalling that for LLSVM the loss augmented decoding subproblem is expressed as
\begin{equation}
\label{eq:hingeloss}
\begin{aligned}
\tilde{H}_k({\bf w}) &= \max_{{\bf y},\mathcal{Z}} H_k({\bf y}, \mathcal{Z}; {\bf w})\\
& = \max_{{\bf y},\mathcal{Z}} \Delta_k({\bf y}, \mathcal{Z}) - \langle {\bf w}, \psi_k({\bf y}, \mathcal{Z})\rangle,
\end{aligned}
\end{equation}
and omitting the lagrangian dual derivation as it is a simple mathematical procedure, we obtain $A_{d\times m}$ to be composed of a set of $m$ columns:
\begin{equation}
A(i, \cdot) = \left\{\frac{\psi_k({\bf y}, \mathcal{Z})^{(i)}}{\lambda K}, \forall k \in \mathbb{N}_K, {\bf y}\in\mathcal{Y}_k,\mathcal{Z}\in{\bf Z}_k\right\},
\end{equation}
where $i$ indicates the row index. Analogously, the column vector ${\bf b}_{m\times1}$ is built as follows.
\begin{equation}
{\bf b} = \left\{\frac{L_k({\bf y}, \mathcal{Z})}{K}, \forall k \in \mathbb{N}_K, {\bf y}\in\mathcal{Y}_k,\mathcal{Z}\in{\bf Z}_k\right\}
\end{equation}
The function $f$ is now differentiated by
\begin{equation}
\begin{aligned}
\nabla f(\boldsymbol\alpha) &= \lambda A^T A\boldsymbol\alpha-{\bf b}^T\\
&=\lambda A^T{\bf w} - {\bf b}^T,
\end{aligned}
\end{equation}
where ${\bf w} = A\boldsymbol\alpha$ is the stationarity KKT condition that has to hold in order to make the duality strong. By substituting the definition of $A$ and ${\bf b}$ for specific values of $(k,{\bf y},\mathcal{Z})$ we obtain
\begin{equation}
\begin{aligned}
\nabla f(\boldsymbol\alpha)_{k,{\bf y},\mathcal{Z}} &= \frac{\lambda}{\lambda K}\psi_k({\bf y}, \mathcal{Z})^T{\bf w} - \frac{1}{K} \Delta_k({\bf y}, \mathcal{Z})\\
&=-\frac{1}{K}\left[\Delta_k({\bf y}, \mathcal{Z}) - \psi_k({\bf y}, \mathcal{Z})^T{\bf w}\right]\\
&=-\frac{1}{K}H_k({\bf y}, \mathcal{Z}).
\end{aligned}
\end{equation}
Which is the same hinge loss of Eq.~\eqref{eq:hingeloss}. So once again the intuition that the linear subproblem that the Frank-Wolfe algorithm has to solve is strongly connected to the loss augmented decoding subproblem is true. Nevertheless, as opposite to the non-latent case, here the hinge loss is also dependent on the latent variable $\mathcal{Z}$ which makes the problem non convex. Thus, before computing the $H_k({\bf y}, \mathcal{Z})$, a latent completion step (line 4 of Alg. 1) is needed in order to ensure $f$ to be continuously differentiable over all the domain except for a finite number of points (sufficient condition). Once we attend these precautions, the latent formulation reduces to the standard SSVM case, and as such, all convergence results also apply to the latent case.

\section*{Appendix B: Computational complexity details}
In this section we provide some details on the computational complexity results we presented in the paper. 
Recall that if the problem of data association is tackled as a whole, the complexity of finding a perfect minimum matching amounts to
\begin{equation}
\mathcal{O}((N+N_o)^3),
\label{eq:comp1}
\end{equation}
according to widely used Hungarian implementations, where $N$ is the number of currently active tracks and detections and $N_o$ is the number of occluded tracks that can still be considered for association.

When considering our method, we have to consider two distinct contributions to the complexity of the overall algorithm:
\begin{itemize}
\item \emph{divide} which is accomplished through a correlation clustering (CC) step on $N$ elements.
\item \emph{conquer} or the application of the Hungarian to the generated sub-problems.
\end{itemize}

The division step can be linear or even sublinear with respect to the number of elements to be splitted $N$. This can be obtained through the many recent approximate solutions of the CC (which is NP-hard optimally), \eg~\cite{a_fast_alg,Demaine2006172}. The conquer step has to be evaluated considering that complex and simple zones are solved differently. Simple zones result in independent subproblems that can be solved directly through the Hungarian. Conversely, the complex zones are not independent and occluded targets have to be considered as well during the overall association subproblem.

Let $K$ be the number of clusters created by the CC and suppose a uniform partition of tracks and detections among these clusters. To simplify the notation, let us call $\alpha = \frac{N}{K}$ the average number of elements (tracks and detections) inside a zone.
If $N$ is the overall number of active tracks and detections, each cluster has approximately $\frac{N}{2K}$ tracks and $\frac{N}{2K}$ detections that need to be associated.  Independently of the zone, these two quantities must coincide for the zone to be simple, so the complexity of solving a simple zone is $\mathcal{O}((\frac{N}{2K})^3) = \mathcal{O}((\frac{\alpha}{2})^3)$. Note that for large $N$, $\alpha$ will typically be much smaller than $N$. If simple clusters are a fraction $\bar{\beta}K$ of the overall number of clusters $K$, the final complexity of solving simple zones is $\frac{\bar{\beta}}{\alpha}N\mathcal{O}((\frac{\alpha}{2})^3)$ which reduces to $\bar{\beta}N\mathcal{O}((\frac{\alpha}{2})^3)$ in the worst case hypotheses. Note that these subproblem can be solved in parallel and typically $\alpha<<N$ when $N$ is large. This is because $\alpha$, the number of interfering tracks/detections, is limited by the non-maxima suppression response of a detector.

Complex zones have to be solved altogether due to the $N_o$ shared occluded tracks. If $\bar{\beta}K$ is the number of simple groups, the number of complex groups is $(1-\bar{\beta})K$ or $\beta K$ for notation convenience. Since the number of tracks and detections are not equal anymore, the number of rows/columns to consider for each group is $\frac{N}{K}$. We thus obtain a complexity of $\mathcal{O}((\beta K \frac{N}{K})^3)$ and due to the addition of occluded targets, the complexity increases to $\mathcal{O}((\beta N + N_o)^3)$ . Summing up we obtain that the overall complexity of the conquer step is
\begin{equation}
(1-\beta)N\mathcal{O}((\frac{\alpha}{2})^3)+\mathcal{O}((\beta N + N_o)^3).
\end{equation}
Note that when $\beta=0$, \ie all zones are simple, only the first term matters; while if $\beta=1$ than the contribution of the first term vanishes and the second term reduces to a stand hungarian over all the tracks/detections as in Eq.~\eqref{eq:comp1}.

\section*{Appendix C: Kalman filtering}
In order to propagate tracks position over occlusions, we employed a simple Kalman Filter predictor with a constant velocity measurement model. This basically means that while unobserved, tracks keep moving by assuming their velocity will not change over time. More formally, the standard discrete Kalman Filter formulation, when no input is considered, is:
\begin{equation}
\begin{aligned}
&x(k) &= &\quad Ax(k-1) + w(k-1)\\
&z(k) &= &\quad Hx(k) + v(k),
\end{aligned}
\end{equation}
being the first equation the state equation and the second one the measurement equation. Here $x(k)$ represents the state of a track at time $k$, while $z(k)$ its measured position. $H$ is the matrix which relates these two variables, namely relates the state and the measurement. $A$ is called state space matrix and explain how the model should evolve over time by means of its physical intrinsic peculiarities. $v(k)$ and $w(k-1)$ are the measurement and state noise random variables. During occlusions, the observation cannot be directly measured so we need to rely on the second relation $z(k) = Hx(k) + v(k)$, and cannot correct the model. A state $x$ for a track is usually represented by a four dimensional vector containing its position and velocity as follows:
\begin{equation}
x(k) = [x_x(k), x_y(k), \dot{x}_x(k), \dot{x}_y(k)]^T
\end{equation}
To describe a constant velocity linear model, we need to specify $A$ and $H$ as follows:
\begin{equation}
A = \begin{bmatrix}
      1 & 0 & 1 & 0           \\
      0 & 1 & 0 & 1			  \\
      0 & 0 & 1 & 0			  \\
	  0 & 0 & 0 & 1
    \end{bmatrix}\quad\quad
H = \begin{bmatrix}
      1 & 0 & 0 & 0           \\
      0 & 1 & 0 & 0			  \\
    \end{bmatrix}
\end{equation}
By substituting the equations, and ignoring the noises just for the sake of simplicity, we obtain the measurement vector:
\begin{equation}
z(k) = [x_x(k-1)+\dot{x}_x(k-1), x_y(k-1)+\dot{x}_y(k-1)]^T,
\end{equation}
which corresponds exactly to a constant velocity model due to the identity 2x2 submatrix in the lower right corner of $A$.

\section*{Appendix D: Detailed experimental results}
In the paper we had to omit some detail on the experimental results. Due to space limitations we presented the results only averaged over the whole sequence set of the Mildly Crowded Dataset (MCD) and the fixed camera sequences of the MOT Challenge (MOT) benchmark. In Tab.~\ref{tab:results_MCD} and Tab.~\ref{tab:results_MOT} we report per sequence results. In particular, for MCD we also report results for the considered competitors; while for the MOT benchmark results are reported for our method only and we let the reader refer to the benchmark site for competitors results:\\
\noindent \url{http://motchallenge.net/results_detail}.

\begin{table*}[t]
\centering
\begin{tabular}{l|r||c|c|c|c|c|c|c|c}
Sequence & Method &  onl. & app. &  MOTA & MOTP & ~~MT~~ & ~~ML~~ & ~~IDS~~ & ~~FRG~~\\
\hline 
\rowcolor{Gray}
{\bf PETS09-S2-L2~~~~} & LDCT (our) & \checkmark & \emph{w.n.}& 								{\bf 47.4} & {\bf 70.8} & 6 & {\bf 3} & 297 & 300\\
\rowcolor{Gray2}
 {\small 42 pedestrian} & LDCT (all features) & \checkmark & \checkmark & 						41.3 & 69.7 & 4 & 7 & 411 & 252\\
 \rowcolor{Gray2}
 {\small up to 33 for frame} & LDCT (only simple) & \checkmark &  & 									35.7 & 68.8 & 3 & 9 & 497 & 323\\
  & Bae and Yun~\cite{bae_robust_2014} & \checkmark & \checkmark & 								30.2 & 69.2 & 1 & 8 & 284 & 499\\
  & Possegger \emph{et al.}~\cite{possegger_occlusion_2014} & \checkmark & & 					40.0 & 68.6 & 8 & 3 & 211 & 342		   \\
 & Milan \emph{et al.}~\cite{milan_continuous_2014} &  & & 										44.9 & 70.2 & 5 & 6 & 150 & 165\\
\hline
\rowcolor{Gray}
{\bf PETS09-S2-L3} & LDCT (our) & \checkmark & \emph{w.n.}& 								 	{\bf 35.2} & {\bf 66.7} & 6 & 15 & 120 & {\bf 12}\\
\rowcolor{Gray2}
 {\small 44 pedestrian} & LDCT (all features) & \checkmark & \checkmark & 						30.6 & 65.1 & 1 & 20 & 235 & 45\\
 \rowcolor{Gray2}
 {\small up to 42 for frame} & LDCT (only simple) & \checkmark &  & 									26.1 & 63.2 & 1 & 25 & 316 & 62\\
  & Bae and Yun~\cite{bae_robust_2014} & \checkmark & \checkmark & 								28.8 & 62.3 & 8 & 17 & 96 & 150\\
  & Possegger \emph{et al.}~\cite{possegger_occlusion_2014} & \checkmark & & 					32.2 & 64.1 & 5 & 12 & 79 & 111 \\
 & Milan \emph{et al.}~\cite{milan_continuous_2014} &  & & 										31.3 & 64.6 & 7 & 23 & 71 & 56 \\
\hline
%
\rowcolor{Gray}					
{\bf GVEII} & LDCT (our) & \checkmark & \emph{w.n.}& 											{\bf 65.6} & {\bf 73.5} & {\bf 208} & {\bf 63} & {\bf 285} & {\bf 71}  \\
\rowcolor{Gray2}
{\small 630 pedestrian} & LDCT (all features) & \checkmark & \checkmark & 						55.6 & 70.5 & 172 & 101 & 548 & 320 \\
 \rowcolor{Gray2}
{\small up to 107 for frame} & LDCT (only simple) & \checkmark &  & 									50.9 & 67.9 & 151 & 113 & 753 & 418\\
 & Bae and Yun~\cite{bae_robust_2014} & \checkmark & \checkmark & 								57.9 & 71.1 & 200 & 75 & 1023 & 320 \\
 & Possegger \emph{et al.}~\cite{possegger_occlusion_2014} & \checkmark & & 					51.1 & 69.8 & 153 & 98 & 844 & 652  \\
 & Milan \emph{et al.}~\cite{milan_continuous_2014} &  & & 										49.3 & 71.2 & 147 & 87 & 312 & 244  \\
 \hline
\rowcolor{Gray}
{\bf 1shatian3} & LDCT (our) & \checkmark & \emph{w.n.}& 										{\bf 42.6} & {\bf 61.0} & 133 & {\bf 23} & {\bf 137} & {\bf 32}\\
\rowcolor{Gray2}
{\small 239 pedestrian} & LDCT (all features) & \checkmark & \checkmark & 						34.7 & 59.9 & 68 & 45 & 592 & 154\\
 \rowcolor{Gray2}
 {\small up to 227 for frame} & LDCT (only simple) & \checkmark &  & 									32.8 & 58.9 & 79 & 52 & 776 & 301\\
 & Bae and Yun~\cite{bae_robust_2014} & \checkmark & \checkmark & 								38.9 & 60.7 & 150 & 40 & 1146 & 185\\
 & Possegger \emph{et al.}~\cite{possegger_occlusion_2014} & \checkmark & & 					31.5 & 57.4 & 110 & 35 & 686 & 654\\
 & Milan \emph{et al.}~\cite{milan_continuous_2014} &  & & 										36.8 & 60.8 & 98 & 50 & 435 & 98\\
\end{tabular}
\caption{Comparison of the proposed method (dark grey) with the state of the art methods on the MCD dataset. In the appearance column, \emph{w.n.} means \emph{when needed}. For each sequence, we also run our code by always associating based on the whole feature set and simple features only (light grey baselines).}
\label{tab:results_MCD}
\end{table*}

\begin{table*}[t]
\centering
\begin{tabular}{l||c|c|c|c|c|c|c|c|c}
Sequence & MOTA (\%) & MOTP (\%) & GT & MT & ML & FP & FN & IDS & FRG\\
\hline 
\rowcolor{Gray}
{\bf TUD-Crossing} & 67.7 & 82.9 & 13 & 9 & 1 & 100 & 205 & 51 & 30\\
\rowcolor{Gray2}
{\bf PETS09-S2L2} & 47.4 & 70.8 & 42 & 6 & 3 & 995 & 3779 & 297 & 300\\
\rowcolor{Gray}
{\bf AVG-TownCentre}\quad\quad\quad\quad\quad & 31.7 & 72.2 & 226 & 36 & 24 & 1878 & 2608 & 395 & 509\\
\rowcolor{Gray2}
{\bf ADL-Rundle-3} & 25.2 & 73.4 & 44 & 2 & 27 & 453 & 7039 & 110 & 101\\
\rowcolor{Gray}
{\bf KITTI-16} & 53 & 79 & 17 & 2 & 3 & 91 & 665 & 44 & 57\\
\rowcolor{Gray2}
{\bf Venice-1} & 33.5 & 68.4 & 17 & 0 & 5 & 578 & 2384 & 73 & 129\\
\hline
\hline
Mean scores & 43.1 & 74.5 & 59.8 & 9.2 & 10.5 & 682.5 & 2780 & 161.7 & 187.7\\
\end{tabular}
\caption{Per sequence results of our method on the MOT Challenge fixed camera sequences. Last row contains mean values and is the one reported in the paper for comparison. Refer to the benchmark website (\url{http://motchallenge.net/results_detail}) for competitors detailed results.}
\label{tab:results_MOT}
\end{table*}

{\small
\bibliographystyle{ieee}
\bibliography{egbib}

\begin{thebibliography}{10}\itemsep=-1pt

\bibitem{alvarez_how_2007}
G.~A. Alvarez and S.~L. Franconeri.
\newblock How many objects can you track?: Evidence for a resource-limited
  attentive tracking mechanism.
\newblock {\em Journal of Vision}, 7(13), Oct. 2007.

\bibitem{bae_robust_2014}
S.-H. Bae and K.-J. Yoon.
\newblock Robust online multi-object tracking based on tracklet confidence and
  online discriminative appearance learning.
\newblock In {\em 2014 {IEEE} Conference on Computer Vision and Pattern
  Recognition ({CVPR})}, pages 1218--1225, June 2014.

\bibitem{bansal_correlation_2002}
N.~Bansal, A.~Blum, and S.~Chawla.
\newblock Correlation clustering.
\newblock {\em Machine Learning}, 56:89--113, July 2004.

\bibitem{Benenson2014Eccvw}
R.~Benenson, M.~Omran, J.~Hosang, and B.~Schiele.
\newblock Ten years of pedestrian detection, what have we learned?
\newblock In {\em European Conference on Computer Vision Workshops}, pages
  613--627. 2015.

\bibitem{berclaz_multiple_2011}
J.~Berclaz, F.~Fleuret, E.~Turetken, and P.~Fua.
\newblock Multiple object tracking using k-shortest paths optimization.
\newblock {\em {IEEE} Transactions on Pattern Analysis and Machine
  Intelligence}, 33(9):1806--1819, Sept. 2011.

\bibitem{bernardin2008evaluating}
K.~Bernardin and R.~Stiefelhagen.
\newblock Evaluating multiple object tracking performance: The clear mot
  metrics.
\newblock {\em EURASIP Journal on Image and Video Processing}, 2008(1):246309,
  2008.

\bibitem{5459278}
M.~Breitenstein, F.~Reichlin, B.~Leibe, E.~Koller-Meier, and L.~Van~Gool.
\newblock Robust tracking-by-detection using a detector confidence particle
  filter.
\newblock In {\em Computer Vision, 2009 IEEE 12th International Conference on},
  pages 1515--1522, Sept 2009.

\bibitem{Demaine2006172}
E.~D. Demaine, D.~Emanuel, A.~Fiat, and N.~Immorlica.
\newblock Correlation clustering in general weighted graphs.
\newblock {\em Theoretical Computer Science}, 361(2–3):172 -- 187, 2006.
\newblock Approximation and Online Algorithms.

\bibitem{dicle_way_2013}
C.~Dicle, O.~Camps, and M.~Sznaier.
\newblock The way they move: Tracking multiple targets with similar appearance.
\newblock In {\em 2013 {IEEE} International Conference on Computer Vision
  ({ICCV})}, pages 2304--2311, Dec. 2013.

\bibitem{dollar_fast_2014}
P.~Dollar, R.~Appel, S.~Belongie, and P.~Perona.
\newblock Fast feature pyramids for object detection.
\newblock {\em {IEEE} Transactions on Pattern Analysis and Machine
  Intelligence}, 36(8):1532--1545, Aug. 2014.

\bibitem{gong2012}
D.~Gong, X.~Zhao, and G.~G. Medioni.
\newblock Robust multiple manifold structure learning.
\newblock In {\em ICML}, 2012.

\bibitem{Goodale199220}
M.~A. Goodale and A.~Milner.
\newblock Separate visual pathways for perception and action.
\newblock {\em Trends in Neurosciences}, 15(1):20 -- 25, 1992.

\bibitem{hofmann_unified_2013}
M.~Hofmann, M.~Haag, and G.~Rigoll.
\newblock Unified hierarchical multi-object tracking using global data
  association.
\newblock In {\em 2013 {IEEE} International Workshop on Performance Evaluation
  of Tracking and Surveillance ({PETS})}, pages 22--28, Jan. 2013.

\bibitem{kahneman_reviewing_1992}
D.~Kahneman, A.~Treisman, and B.~J. Gibbs.
\newblock The reviewing of object files: Object-specific integration of
  information.
\newblock {\em Cognitive Psychology}, 24(2):175--219, Apr. 1992.

\bibitem{kuhn_hungarian_1955}
H.~W. Kuhn.
\newblock The hungarian method for the assignment problem.
\newblock {\em Naval Research Logistics Quarterly}, 2(1-2):83--97, Mar. 1955.

\bibitem{lacoste-julien_block-coordinate_2013}
S.~Lacoste-Julien, M.~Jaggi, M.~Schmidt, and P.~Pletscher.
\newblock Block-coordinate frank-wolfe optimization for structural {SVMs}.
\newblock In {\em International Conference on Machine Learning}, 2013.

\bibitem{a_fast_alg}
J.~Li, X.~Huang, C.~Selke, and J.~Yong.
\newblock A fast algorithm for finding correlation clusters in noise data.
\newblock In Z.-H. Zhou, H.~Li, and Q.~Yang, editors, {\em Advances in
  Knowledge Discovery and Data Mining}, volume 4426 of {\em Lecture Notes in
  Computer Science}, pages 639--647. Springer Berlin Heidelberg, 2007.

\bibitem{5206735}
Y.~Li, C.~Huang, and R.~Nevatia.
\newblock Learning to associate: Hybridboosted multi-target tracker for crowded
  scene.
\newblock In {\em Computer Vision and Pattern Recognition, 2009. CVPR 2009.
  IEEE Conference on}, pages 2953--2960, June 2009.

\bibitem{milan_continuous_2014}
A.~Milan, S.~Roth, and K.~Schindler.
\newblock Continuous energy minimization for multitarget tracking.
\newblock {\em {IEEE} Transactions on Pattern Analysis and Machine
  Intelligence}, 36(1):58--72, Jan. 2014.

\bibitem{motchallenge}
L.~Milan, A. Leal-Taixé, K.~Schindler, S.~Roth, and I.~Reid.
\newblock {MOT} {C}hallenge.
\newblock \url{http://www.motchallenge.net}, 2014.

\bibitem{possegger_occlusion_2014}
H.~Possegger, T.~Mauthner, P.~M. Roth, and H.~Bischof.
\newblock Occlusion geodesics for online multi-object tracking.
\newblock In {\em 2014 {IEEE} Conference on Computer Vision and Pattern
  Recognition ({CVPR})}, pages 1306--1313, June 2014.

\bibitem{pylyshyn_role_1989}
Z.~Pylyshyn.
\newblock The role of location indexes in spatial perception: a sketch of the
  {FINST} spatial-index model.
\newblock {\em Cognition}, 32(1):65--97, June 1989.

\bibitem{smeulders_visual_2014}
A.~Smeulders, D.~Chu, R.~Cucchiara, S.~Calderara, A.~Dehghan, and M.~Shah.
\newblock Visual tracking: An experimental survey.
\newblock {\em {IEEE} Transactions on Pattern Analysis and Machine
  Intelligence}, 36(7):1442--1468, July 2014.

\bibitem{wu_online_2013}
Z.~Wu, J.~Zhang, and M.~Betke.
\newblock Online motion agreement tracking.
\newblock In {\em Proceedings of the British Machine Vision Conference}. BMVA
  Press, 2013.

\bibitem{Yang_2014}
B.~Yang and R.~Nevatia.
\newblock Multi-target tracking by online learning a crf model of appearance
  and motion patterns.
\newblock {\em Int. J. Comput. Vision}, 107(2):203--217, Apr. 2014.

\bibitem{yu_learning_2009}
C.-N.~J. Yu and T.~Joachims.
\newblock Learning structural {SVMs} with latent variables.
\newblock In {\em Proceedings of the 26th Annual International Conference on
  Machine Learning}, {ICML} '09, pages 1169--1176, New York, {NY}, {USA}, 2009.
  {ACM}.

\bibitem{6714561}
B.~Zhou, X.~Tang, H.~Zhang, and X.~Wang.
\newblock Measuring crowd collectiveness.
\newblock {\em Pattern Analysis and Machine Intelligence, IEEE Transactions
  on}, 36(8):1586--1599, Aug 2014.

\end{thebibliography}
}

\end{document}